\documentclass[acmtog,dvipsnames,screen=true,nonacm]{acmart}  

\usepackage{tabularx,booktabs} 
\usepackage[normalem]{ulem}  
\usepackage{hyphenat,balance}
\usepackage[fleqn,tbtags]{mathtools}
\usepackage{overpic}  
\usepackage{wrapfig}
\usepackage{contour}
\usepackage{microtype}
\usepackage{multirow}
\usepackage{colortbl}
\usepackage{amsmath}
\usepackage{bbm}
\usepackage{textcomp}
\usepackage{enumitem}
\setitemize{noitemsep,topsep=0pt,parsep=0pt,partopsep=0pt}
\usepackage{derivative}
\usepackage{float}
\usepackage{graphicx}
\graphicspath{{./images/}}

\usepackage{color_style}
\usepackage{hyperref} 
\usepackage{cleveref} 

\usepackage{listings}
\usepackage[ruled]{algorithm2e} 

\SetAlFnt{\small}
\SetAlCapFnt{\small}
\SetAlCapNameFnt{\small}
\SetAlCapHSkip{0pt}

\def \eg {{\emph{e.g.},\thinspace}}
\def \ie {{\emph{i.e.},\thinspace}}

\newcommand{\mypar}[1]{\vspace{1mm}\noindent\textbf{#1}}

\newcommand{\mname}{\emph{ScaleBlind}\xspace}

\begin{document}

\title{\mname: Point Cloud Completion under Unknown Scale}

\author{Shenghui Wu}
\affiliation{\institution{Shandong University} 
\country{China}}
\email{shenghui.wu@mail.sdu.edu.cn}

\author{Chen Wang}
\affiliation{\institution{Shandong University} 
\country{China}}
\email{chen.wang@mail.sdu.edu.cn}

\author{Yuan Feng}
\affiliation{\institution{Shandong University} 
\country{China}}
\email{yfeng@mail.sdu.edu.cn}

\author{Guangshun Wei}
\authornote{Corresponding authors: Guangshun Wei and Yuanfeng Zhou.}
\affiliation{  \institution{Shandong University}
\country{China}}
\email{guangshunwei@gmail.com}

\author{Yuanfeng Zhou}
\authornotemark[1]
\affiliation{\institution{Shandong University} 
\country{China}}
\email{yfzhou@sdu.edu.cn}

\author{Changjian Li}
\affiliation{\institution{University of Edinburgh} 
\country{United Kingdom}}
\email{Changjian.li@ed.ac.uk}

\begin{teaserfigure}
    \centering
    \begin{overpic}[width=0.99\textwidth]{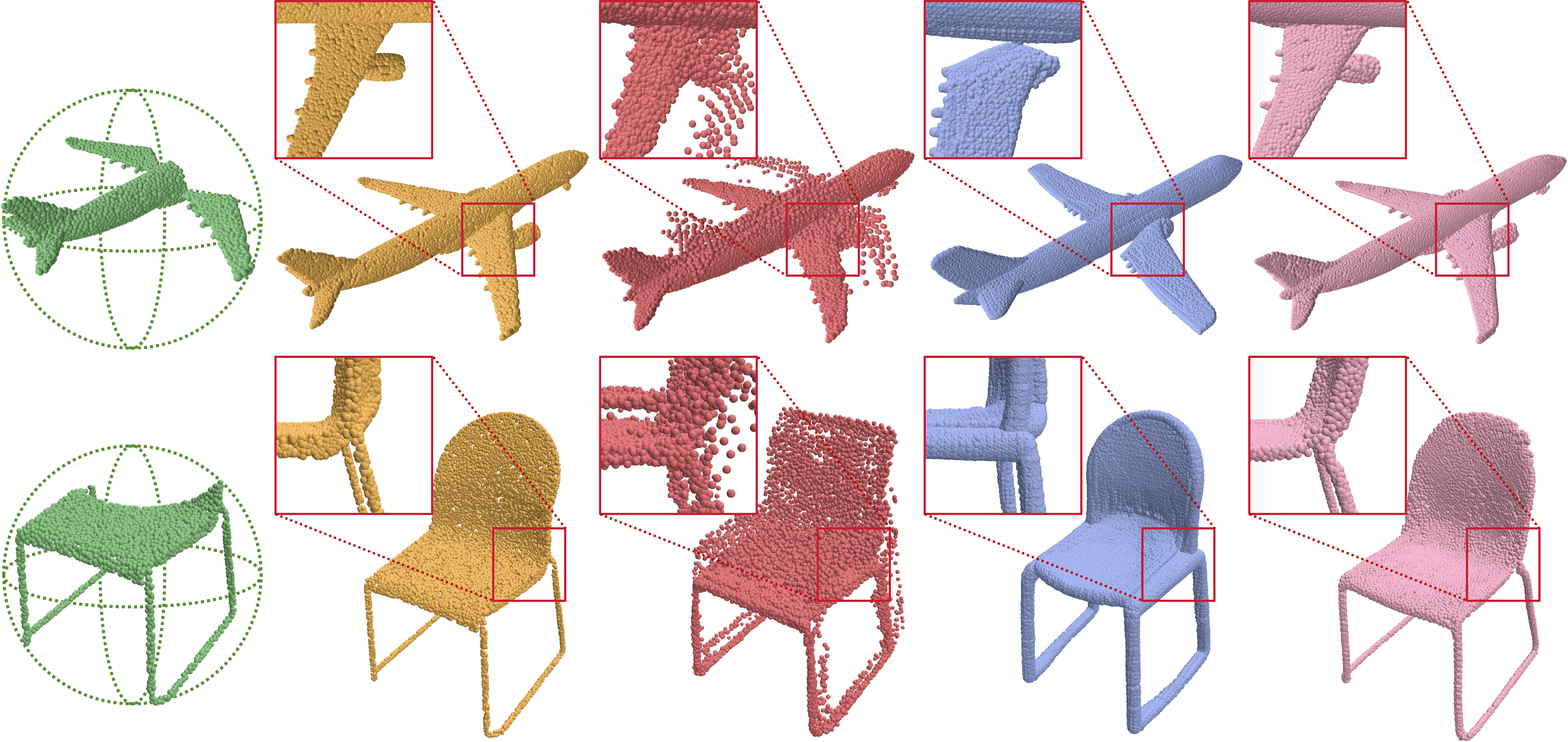} 
    \put(1.5,-1.5) {\small (a) Partial point cloud}
    \put(21,-1.5) {\small (b) Ground truth}
    \put(40,-1.5) {\small (c) GeoFormer~\shortcite{yu2024geoformer}}
    \put(63,-1.5) {\small (d) Trellis~\shortcite{Xiang_2025_CVPR}}
    \put(81,-1.5) {\small (e) \mname (Ours)}
    \end{overpic}
    \caption{\textbf{Point cloud completion.} 
    Given (a) a partial point cloud, supervised completion methods, such as (c) GeoFormer~\cite{yu2024geoformer}, often implicitly rely on access to the ground truth shape scale during inference, leading to over-completion artifacts when such information is unavailable. 
    In contrast, (d) Trellis~\cite{Xiang_2025_CVPR}, a modern 3D generative model, generates structurally plausible complete shapes, yet fails to align with the partial input.
    We propose (e) \mname, a GT-scale-free framework that faithfully completes the shape while strictly adhering to the partial observation.
    }
    \label{fig:teaser}
\end{teaserfigure}
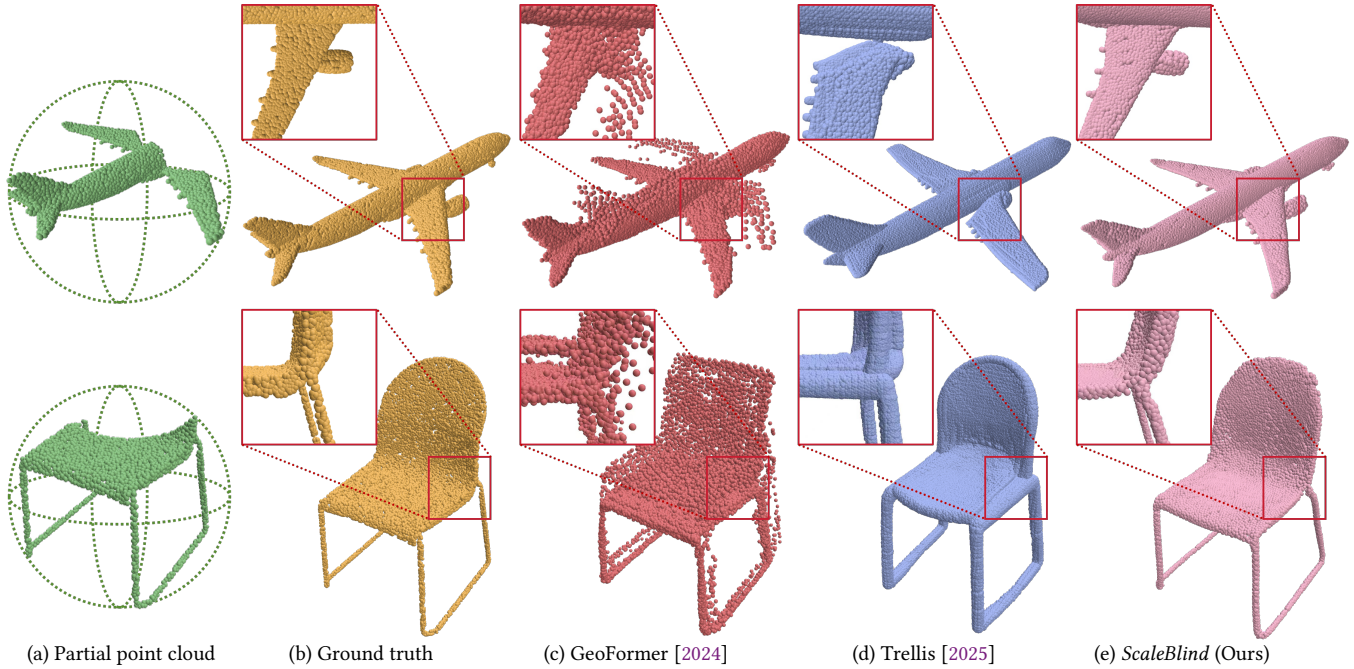

\begin{abstract}

Point cloud completion aims to infer a complete 3D shape from a partial point cloud and serves as a fundamental building block for downstream tasks such as reconstruction, editing, and simulation.
Despite the recent progress, existing learning-based methods often implicitly rely on access to the ground-truth shape scale (GT-scale) during both training- and testing-time normalization, assuming privileged information that is unavailable in real-world inference. 
This hidden assumption limits practical deployment and can lead to severe completion artifacts, \eg over- or under-completion and nested shells, once the oracle GT-scale cue is removed.
We observe that the recent foundation image generation models exhibit a strong capability of understanding objects and geometries, and producing multi-view consistent renderings, making them promising priors for GT-scale-free 3D completion.
Motivated by this insight, we propose \mname, a novel framework that leverages foundation-model-based image completion to recover global scale directly from partial inputs and then faithfully produces the 3D completion.
Specifically, \mname dreams out complete multi-view appearances from rendered partial views, lifts the inferred missing regions back into 3D to obtain a geometry-aware coarse completion, and further refines it via a powerful cross-modal fusion network with the original partial point cloud.
By harnessing 2D foundation priors, our method eliminates the need for accessing GT-scale information at inference. 
Moreover, it provides a principled bridge between 2D generative priors and 3D point cloud completion.
Extensive experiments demonstrate the superiority of our framework, making \mname the new state-of-the-art for the point cloud completion task.


\end{abstract}

%
%


\maketitle

\section{Introduction}

Point cloud completion, \ie recovering a complete 3D shape from a partial point cloud, is a long-standing problem with broad impact in graphics and vision.
It is a critical component for downstream applications such as reconstruction pipelines~\cite{yang2025monge,huang2023nksr}, robot manipulation~\cite{varley2017shape}, and AR/VR content creation~\cite{guo20223d}, where partial observations are the norm. 
However, real-world point clouds acquired from commodity depth sensors or LiDAR are frequently incomplete due to occlusion, limited viewpoints, and sparse sampling, making completion both practically necessary and intrinsically ambiguous. 

\begin{figure}[!t]
    \centering
    \begin{overpic}[width=0.95\linewidth]{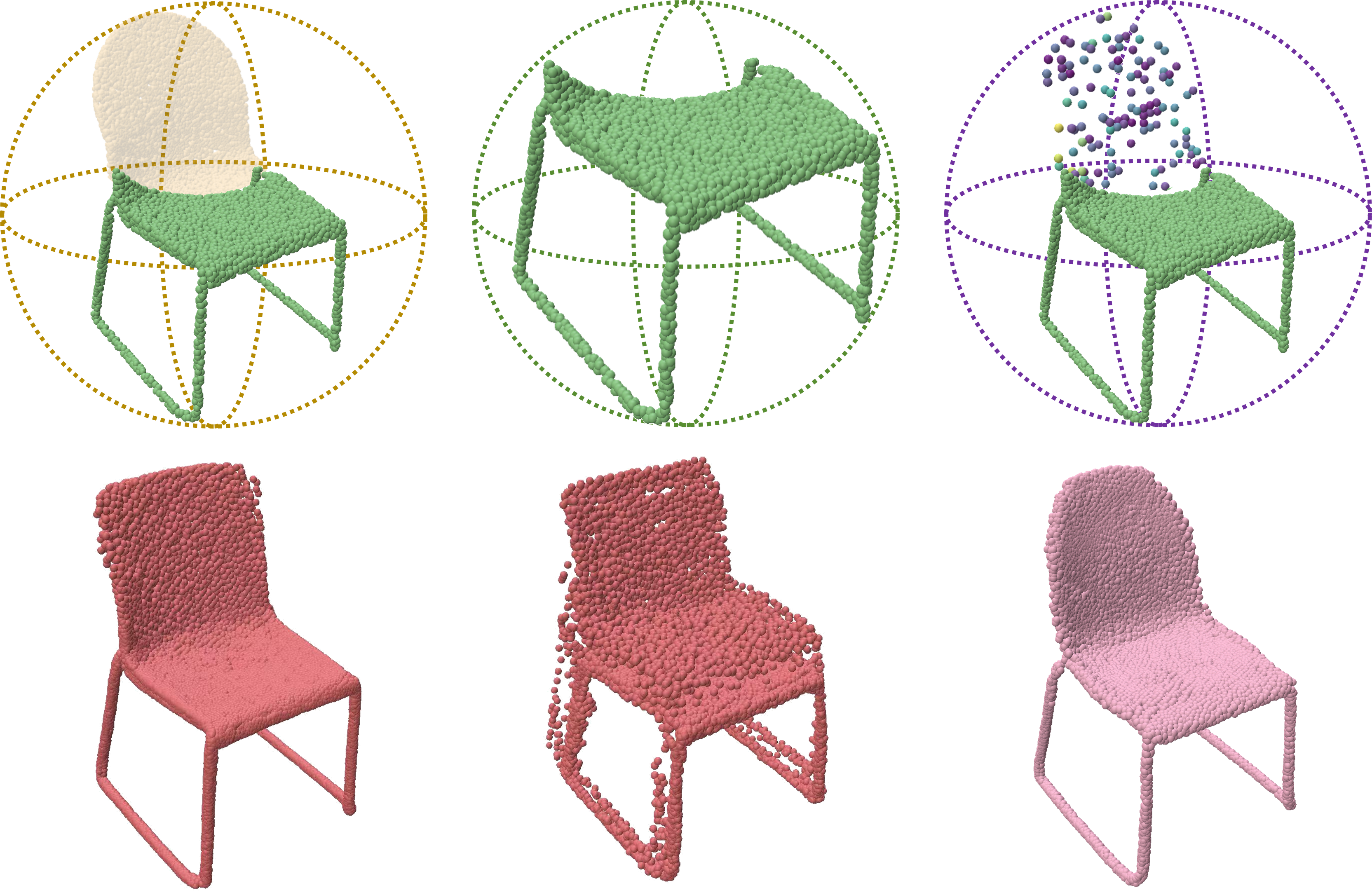} 
    \put(5, -3) {\small (a) W/ GT-scale}
    \put(40,-3) {\small (b) W/o GT-scale}
    \put(80,-3) {\small (c) Ours}
    \end{overpic}
    \caption{\textbf{GT shape scale and completion.} 
    Given a partial point cloud, (a) existing learning-based methods (\eg GeoFormer~\cite{yu2024geoformer}) often assume access to the ground-truth shape scale during testing-time normalization (\eg the missing yellow back part), which provides privileged oracle information and leads to seemingly satisfactory completions. 
    (b) In realistic settings where such information is unavailable in the normalization step, the resulting completions exhibit clear artifacts, such as over-completion and nested shells.
    (c) Motivated by this observation, we propose a novel solution to \emph{`Dream'} the missing scale information at inference (\ie the back part), enabling faithful and scale-consistent completion without GT access.}
    \label{fig:motivation}
    \vspace{-3mm}
\end{figure}

A large body of learning-based methods tackles completion by learning a direct mapping from partial inputs to full shapes.
They extract global and local features and decode missing geometry in a coarse-to-fine manner.
Beyond the ambiguity caused by severe occlusions, we observe a protocol issue that can simplify the task in synthetic benchmarks. 
That is, in many pipelines~\cite{yuan2018pcn,chen2023anchorformer,xiang2021snowflakenet,zhou2022seedformer,wen2022pmp}, shapes are canonicalized (\eg, centered and scaled into a unit sphere) \emph{before} partial observations are generated.
This practice implicitly ties the partial input to the ground-truth global scale and placement (see Fig.~\ref{fig:motivation}(a)), introducing a privileged cue that is not accessible at test time in practical scenarios.
When this oracle scale cue is removed (\ie GT-scale-independent preprocessing, Fig.~\ref{fig:motivation}(b)), the performance of existing completion models can deteriorate sharply, often manifesting as global-scale distortions and over-completion artifacts.

In parallel, recent advances in 3D generative modeling enable high-quality shape synthesis from image or text prompts~\cite{Xiang_2025_CVPR,wu2025directds}.
A natural idea is to render a partial point cloud into a set of views, complete the partial renderings using strong pretrained image models~\cite{li2024craftsman,liu2024oneplus}, and then reconstruct a full 3D shape.
However, these generative systems are not designed for completion: their outputs are not explicitly constrained to remain consistent with the observed partial geometry.
As illustrated in Fig.~\ref{fig:teaser}(d), the wing region of the airplane contains irrational connection structures and suffers from the loss of partial geometric details.
Moreover, the generated model exhibits a certain degree of scale deviation.
For instance, the overall thickness of the chair (Fig.~\ref{fig:teaser}(d)) differs from that of the original model.
Such deviations are particularly problematic in the unknown-scale setting, where scale and placement must be recovered rather than assumed.

Motivated by these observations, we propose \mname, a GT-scale-free point cloud completion framework designed for inference without access to ground-truth scale information.
Our key insight is that recent foundational image generation models possess strong geometric understanding and can reason out missing object evidence with high multi-view consistency, providing a powerful prior for recovering the global extent of a shape from partial geometry.
Specifically, \mname renders the partial point cloud from multiple viewpoints and completes the resulting multiple renderings using a pre-trained large VLM (\ie Nano-banana~\cite{team2023gemini}), yielding plausible full-view appearances and missing regions. 
We then lift these multi-view completions into 3D via a visual-hull-alike voxel voting scheme, forming a probabilistic volumetric prior. 
This prior is further used to extract high-confidence missing point candidates and to estimate an oriented bounding box (OBB) for joint normalization together with the partial input. 
Finally, the normalized partial points and confidence-weighted candidates are fused and processed by a prior-guided coarse-to-fine completion network, where high-level multi-view features first guide the generation of a coarse geometric scaffold, followed by two refinement stages that progressively inject mid-/low-level image cues to recover detailed structures and geometry, producing the final completion while preserving the observed geometry.

Crucially, our framework excels on two parallel fronts: (i) establishing a state-of-the-art 3D completion architecture, and (ii) addressing the practical GT-scale-free challenge. 
This robust geometric capability stems from our novel cascaded refinement network. It first constructs a coarse geometric scaffold guided by global semantics, and then progressively carves out fine-grained details via a powerful dual-mode fusion strategy. 
By simultaneously measuring 3D geometric discrepancies and injecting hierarchical 2D image cues, the network effectively prevents wild hallucinations and faithfully recovers intricate structures. 
Consequently, extensive experiments demonstrate the dual superiority of our approach: under the conventional GT-scale-tied setting, \mname sets a new state-of-the-art, validating the representational power of our core architecture; simultaneously, under the realistic GT-scale-free setting, it exhibits a commanding advantage, producing well-proportioned completions free from severe geometric distortions.

To summarize, our principal contributions are three-fold:
\begin{itemize}
    \item We formulate \emph{GT-scale-free} point cloud completion under unknown global scale, and reveal that GT-scale-tied canonicalization in existing pipelines introduces an unrealistic \emph{scale prior leakage} into partial inputs.
    \item We propose to recover missing global extent by leveraging a pretrained VLM to complete multi-view renderings of the partial scan, and lift the multi-view evidence into a scale-consistent probabilistic 3D prior via voxel voting.
    \item We incorporate the recovered prior into a prior-guided coarse-to-fine completion network with joint normalization and confidence-aware fusion, yielding faithful completions that remain consistent with the observed geometry under GT-scale-independent preprocessing.
\end{itemize}

\begin{figure*}[!t]
    \centering
    \begin{overpic}[width=0.95\linewidth]{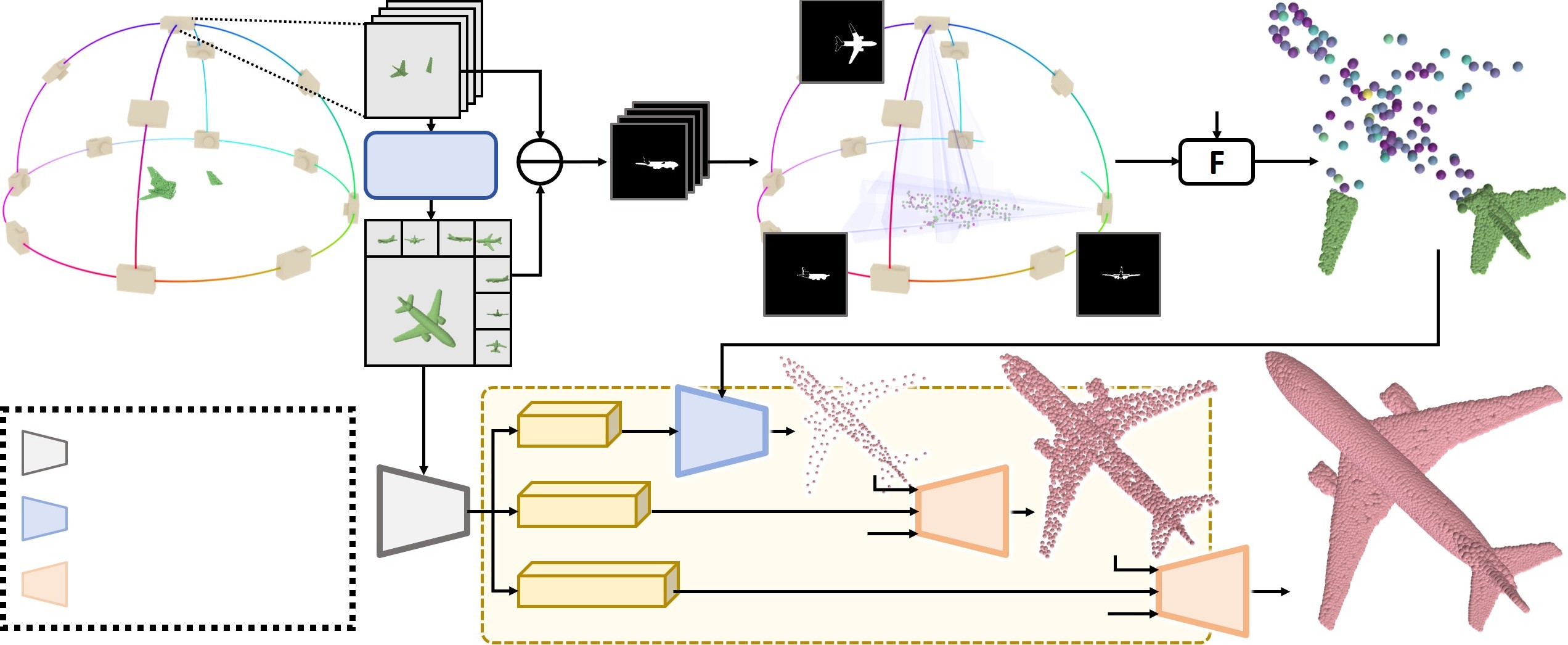} 
    
    \put(4, 20) {\small Multi View Projection}
    \put(9, 26.5) {\small Partial Pointcloud}
    \put(13.5, 24.5) {\small $\mathbf{P}_\text{part}$}
    
    \put(5.5, 11.6) {\small Image Pyramid Encoder}
    \put(5.5, 7.5) {\small Coarse Point Generation}
    \put(5.5, 3.4) {\small Refinement}

    \put(31.2, 39.5){\small Multi-View}
    \put(31.2, 37.7){\small Partial Images}

    \put(25.8, 31){\small \textbf{VLM}}
    \put(23.7, 29.2){\small \textbf{Inpainting}}

    \put(33.2, 20.8){\small Multi-View}
    \put(33.2, 19.0){\small Pred. Images}

    \put(38.3, 27){\small Pixel-wise}
    \put(37.5, 25.2){\small Differencing}
    \put(39.4, 23.4){\small Masks}

    \put(33.2, 11.3){\small $\mathcal{F}_\text{global}$}
    \put(33.2, 6.2){\small $\mathcal{F}_\text{mid}$}
    \put(33.2, 1){\small $\mathcal{F}_\text{low}$}
    
    \put(51.3, 6.8){\small $\mathbf{P}_\text{part}$}
    \put(66.6, 1.8){\small $\mathbf{P}_\text{part}$}

    \put(62.5, 30.9){\small Confidence}
    \put(64.3, 29.1){\small Field}

    \put(75.5, 27.4){\small Fusion}
    
    \put(76.2, 34.8){\small $\mathbf{P}_\text{part}$}

    \put(86, 0) {\small Final Shape}
    \end{overpic}
    \caption{\textbf{Method Overview.} 
    Our method consists of two core modules - prior acquisition (top, Sec.~\ref{subsec:prior_acquisition}) and prior-guided point completion (bottom, Sec.~\ref{subsec:completion}). 
    \emph{Prior acquisition}: given a partial point cloud $\mathbf{P}_\text{part}$ (the gray silhouette reveals the large missing portion), our method first exploits a pretrained VLM to inpaint the partial renders and then obtains the complete shape prior via a visual-hull-like voxel voting.
    \emph{Point completion}: guided by the full shape prior, our method first produces a complete coarse point cloud, and then refines and consolidates the shape in two refinement steps by injecting hierarchical image cues.
    }
    \label{images:pipeline}
    \vspace{-3mm}
\end{figure*}
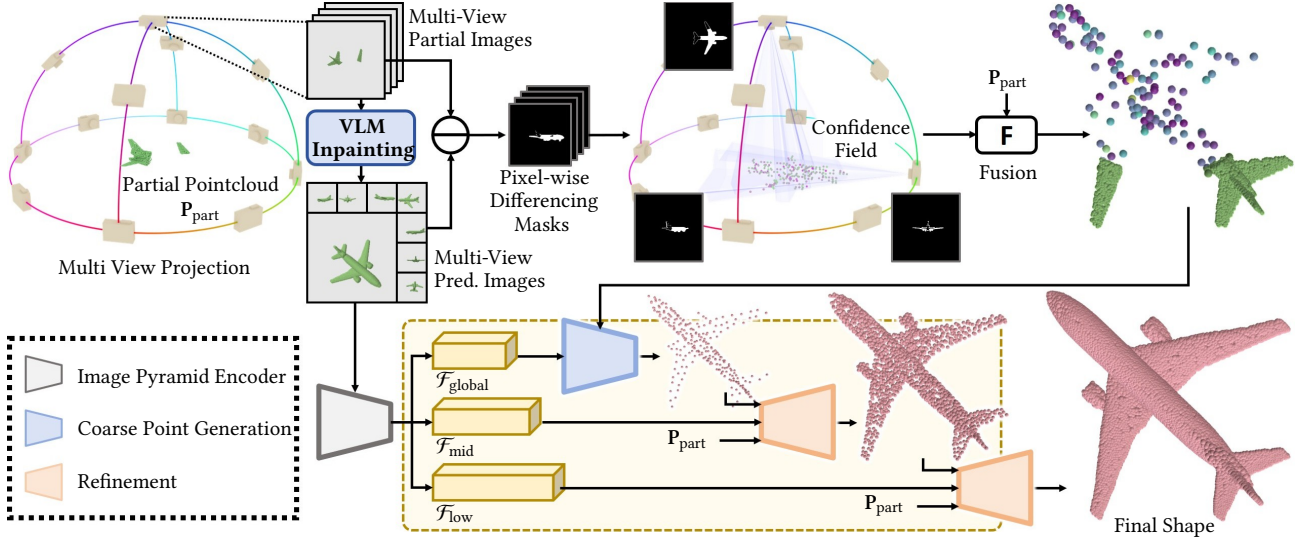

\section{Related Work}

\paragraph{Point-based Learning Methods.}
Deep learning-based point cloud completion~\cite{yu2021pointr,xiang2021snowflakenet,yan2022fbnet,tang2022lake} has emerged as a predominant research direction. PointNet~\cite{qi2017pointnet} established the foundation for directly processing 3D coordinates. Building on this, PCN~\cite{yuan2018pcn} pioneered an end-to-end framework that operates directly on raw point clouds, utilizing a coarse-to-fine strategy to progressively generate complete shapes. Subsequently, Huang et al.~\cite{huang2020pf} further optimized this framework by refining the multi-stage completion process, significantly enhancing overall performance. Most of these methods assume canonicalized inputs during training and evaluation, and do not explicitly address completion under unknown global scale.

\paragraph{Cross-modal Point-Image Fusion.}
To compensate for the limited geometric cues in partial point cloud scans, researchers have incorporated auxiliary modalities—such as RGB or depth images—to be processed alongside the geometric data. For instance, ViPC~\cite{zhang2021view} fused partial point clouds with single-view RGB images to achieve view-guided completion. SVDFormer~\cite{zhu2023svdformer} integrated multi-view depth information with point-based features, utilizing Transformer modules to effectively fuse the spatial relationships of neighboring points with visual semantics. GeoFormer~\cite{yu2024geoformer} further enhanced global geometric expression by aligning image features from multi-view Consistent Canonical Maps (CCM) with point cloud features via attention mechanisms. PCDreamer~\cite{wei2025pcdreamer} further explores multi-view generative priors to guide completion. Despite these advances, a common limitation remains in synthetic benchmarks: the full shape is often canonicalized before partial observations (and any rendered auxiliary views) are generated, which implicitly ties the input to GT-consistent scale. This assumption is problematic in real scans, where the global extent of the missing region is not available at inference.

\paragraph{Vision-Language Models.}
The rise of Large Vision-Language Models (VLMs) has provided a powerful paradigm for structural reasoning and image-level completion. Unlike traditional generative models, VLMs—leveraging architectures similar to LLaVA~\cite{liu2023llava} and other multimodal frameworks—possess extensive pre-trained knowledge that enables zero-shot reasoning of object structures. In 3D tasks, researchers have begun exploring VLMs to interpret partial observations and hypothesize missing components. For instance, Point-Bind~\cite{guo2023pointbindpointllmaligning} and Point-LLM~\cite{xu2024pointllm} demonstrate the potential of aligning 3D geometry with VLM-based semantic spaces. 

Overall, most existing completion methods inherently rely on GT-scale-tied canonicalization, while pure image-to-3D generative models fail to preserve the observed partial geometry and are highly prone to scale drift. In contrast, our framework achieves breakthroughs on two parallel fronts. First, as a fundamental completion architecture, we propose a state-of-the-art prior-guided coarse-to-fine network that meticulously aligns multi-modal features to faithfully recover intricate 3D structures. Second, to fundamentally resolve the realistic GT-scale-free challenge under unknown scale, we leverage a VLM to inpaint multi-view projections and lift them into a probabilistic 3D prior via voxel voting. By performing prior-based joint normalization alongside our robust completion network, we yield scale-consistent and geometry-preserving shapes without relying on any privileged oracle information.
\section{Method}
\label{sec:method}

\subsection{Overview}
\label{subsec:overview}
\begin{figure}[!t]
    \centering
    \begin{overpic}[width=0.95\linewidth]{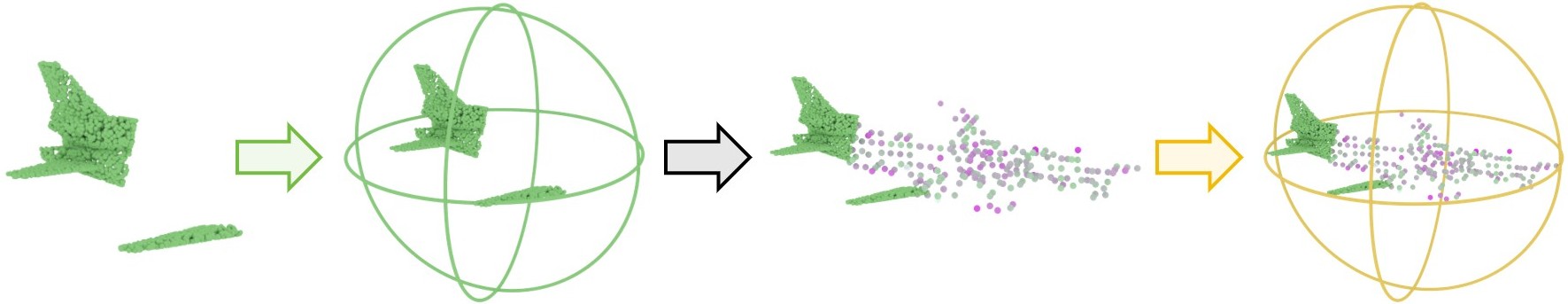} 
    \put(8, 15) {\small $\mathbf{P}_{\text{part}}$}
    \put(10, -1) {\small Partial-only}
    \put(8, -4.5) {\small normalization}
    \put(60, 15) {\small $\mathbf{P}_{\text{merged}}$}
    \put(38, -1) {\small Candidates}
    \put(34, -4.5) {\small voting and merge}
    \put(63, -1) {\small Prior-based joint}
    \put(65, -4.5) {\small normalization}
    \end{overpic}
    \caption{\textbf{Illustration of our normalization strategies.} We first apply a \emph{partial-only normalization} to the input partial point cloud $\mathbf{P}_{\text{part}}$. After extracting and merging the candidate points to form $\mathbf{P}_{\text{merged}}$, a \emph{prior-based joint normalization} is performed to dynamically adjust the scale based on the combined spatial extent of the partial input and the generated geometric priors.}
    \label{fig:normalization}
    \vspace{5mm}
\end{figure}
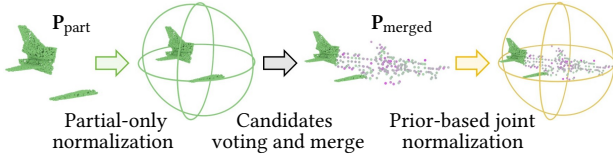
\paragraph{Problem definition.}
Given the partial point cloud input $\mathbf{P}_\text{part}$, the goal of point cloud completion is to produce the full shape $\mathbf{P}_\text{full}$, respecting the partial observation.
In the literature of supervised point cloud completion, a common preprocessing step is to normalize the partial input (\ie centering and scaling into a unit sphere). 
When the missing region is structural-preserving (\eg the middle of a chair leg), the on-the-fly normalization of the partial input produces consistent scale and placement as in the ground truth point cloud.
However, as shown in Fig.~\ref{fig:motivation}(a), when the missing region is structure-disruptive (\eg the chair back part), GT scale information is still incorporated to match the training time operation, so as to produce satisfactory shapes. 
Clearly, this scale information is inaccessible in real-world inference, causing severe performance degradation (Fig.~\ref{fig:motivation}(b)).
We introduce a GT-scale-free point cloud completion task, where the observed partial point cloud is independently normalized, and a prior acquisition module is employed to predict GT-scale cues from the partial observation rather than assuming them.

\paragraph{Method.} 
To solve this new task, we propose \mname, consisting of two core modules - prior acquisition and prior-guided completion. Figure~\ref{images:pipeline} illustrates a method overview, and in what follows, we elaborate on technical details.

\subsection{Prior Acquisition}
\label{subsec:prior_acquisition}

The goal of this module is to estimate the coarse full shape prior from the partial observation in the following four steps.

\paragraph{Multi-view projection and completion.}
As illustrated in Fig.~\ref{images:pipeline}(top), we sample a set of viewpoints $\mathcal{V}$ on the upper hemisphere surrounding the partial input and place the camera at a sufficiently large distance (default $r{=}5$). 
Unlike conventional pipelines that render from a unit sphere, our framework performs partial-only normalization, thus the global extent of the missing geometry is unknown \emph{a priori}. We therefore adopt a conservative camera distance, chosen empirically for extremely high missing ratios, to provide a sufficiently wide field of view that covers the potential completion region. By maintaining this camera distance, we prevent the VLM-hallucinated structures from extending beyond the image boundaries.
Subsequently, we apply Nano-banana~\cite{team2023gemini} to each viewpoint to synthesize completed projections $\{I_{\text{comp},v}\}_{v\in\mathcal{V}}$ that capture the approximate global shape, thereby generating plausible missing regions in the 2D domain. See \emph{supplementary} for the detailed prompt when interacting with the VLM.

To detect the inpainted regions that indicate the missing geometry, we compute a \emph{mask} $M_v$ per view by pixel-wise differencing:
\begin{equation}
M_v(i,j)=
\begin{cases}
1, & \text{if } \left|I_{\text{comp},v}(i,j)-I_{\text{part},v}(i,j)\right|>\delta_{\text{diff}},\\
0, & \text{otherwise},
\end{cases}
\label{eq:hallucination_mask}
\end{equation}
where $\delta_{\text{diff}}$ is a fixed intensity threshold.

\paragraph{Geometric consistency check.}
VLM completion may occasionally introduce view-wise structures and scales that are incompatible with the observed evidence.
We therefore filter unreliable viewpoints using a simple overlap test.
Let $B_{p,v}$ and $B_{c,v}$ denote the binary foreground masks of the partial and completed projections, respectively.
We compute the overlap ratio:
\begin{equation}
{r}_v=\frac{\sum_{(i,j)}\left(B_{p,v}\cap B_{c,v}\right)}{\sum_{(i,j)} B_{p,v}},
\label{eq:overlap_ratio}
\end{equation}
and keep a view in the valid set $\mathcal{V}_{\text{valid}}$ only if ${r}_v\ge \tau_{\text{overlap}}$, which suppresses clearly misaligned hallucinations.

\paragraph{Multi-view voting.}
To get the 3D shape prior, we lift the multi-view completions into 3D using a visual-hull-like voxel voting scheme.
Following the partial-only normalization strategy illustrated in Fig.~\ref{fig:normalization}, we define a canonical voxel grid $\mathcal{G}$ over the input partial point cloud space. This grid is bounded within a cubic coordinate range of $[-5, 5]^3$ and parameterized by a voxel resolution $\Delta$.
For each voxel center $\mathbf{x}\in\mathbb{R}^3$ and each valid view $v$, we project $\mathbf{x}$ onto the image plane:
\begin{equation}
\mathbf{p}_v = \pi\!\left(\mathbf{K}(\mathbf{R}_v\mathbf{x}+\mathbf{t}_v)\right),
\label{eq:projection}
\end{equation}
where $\pi(\cdot)$ is the perspective projection operator, and $\mathbf{K}$, $(\mathbf{R}_v,\mathbf{t}_v)$ are the camera intrinsics and extrinsics.
A voxel $\mathbf{x}$ receives a vote from view $v$ if its projection falls inside the foreground of $M_v$.
The accumulated votes are:
\begin{equation}
S(\mathbf{x}) = \sum_{v\in\mathcal{V}_{\text{valid}}} M_v\!\left(\lfloor \mathbf{p}_v \rfloor\right),
\label{eq:voting}
\end{equation}
where $\lfloor\cdot\rfloor$ maps image coordinates to pixel indices.

We normalize the voting volume into a probabilistic field:
\begin{equation}
\mathrm{P}_{\text{prior}}(\mathbf{x}) = \frac{S(\mathbf{x})}{\max_{\mathbf{x}'\in\mathcal{G}}S(\mathbf{x}')},
\label{eq:prior_field}
\end{equation}
where each voxel value $\mathbf{P}_{\text{prior}}(\mathbf{x})\in[0,1]$ quantifies the consensus confidence of the missing geometry supported by multi-view VLM completions.
This probabilistic field serves as a shape-aware and scale-consistent prior for completion.

\paragraph{Prior-based joint normalization.}
Given the cubic spatial extent of $[-5, 5]^3$ and the voxel resolution $\Delta = 0.02$, the predicted probabilistic field is formulated as a volumetric tensor $\mathrm{P}_{\text{prior}} \in \mathbb{R}^{500 \times 500 \times 500}$. From this tensor, we further extract a concrete high-confidence candidate point set by applying a probability threshold $\tau_{\text{prob}}$.
These candidate points provide a coarse estimation of the missing geometry.
We then jointly normalize the partial point cloud and the candidate set into a unit sphere according to their combined spatial extent, yielding $\tilde{\mathbf{P}}_{\text{part}}$ and $\tilde{\mathbf{P}}_{\text{cand}}$.
Importantly, this normalization does not require any GT-scale information, yet yields a faithful global-scale normalization by recovering the coarse missing extent from multi-view priors.

\subsection{Prior-guided Completion}
\label{subsec:completion}

Given the normalized partial point cloud $\tilde{\mathbf{P}}_{\text{part}}$ and  $\tilde{\mathbf{P}}_{\text{cand}}$, we aim to predict a complete point set that follows the inferred global structure while preserving the observed geometry.
We design the completion module in a coarse-to-fine manner using a bespoke neural network -- it first predicts a coarse scaffold and then refines it in two cascaded refinement steps (see Fig.~\ref{images:pipeline}(bottom)).

\paragraph{Partial and candidate points fusion.}
To initialize the completion process, we first merge the input partial points with the generated prior candidates to form a fused set:
\begin{equation}
\tilde{\mathbf{P}}_{\text{merged}}= \texttt{FPS} \left( \tilde{\mathbf{P}}_{\text{part}}\cup \tilde{\mathbf{P}}_{\text{cand}}, N \right) \in \mathbb{R}^{N \times 4}.
\label{eq:merge}
\end{equation}

Other than the coordinate, each fused point has an extra confidence attribute $\sigma$.
We set $\sigma=1.0$ for input partial points, while querying $\mathrm{P}_{\text{prior}}$ for points in $\tilde{\mathbf{P}}_{\text{cand}}$ at their respective locations.
Finally, we apply farthest point sampling (\texttt{FPS}) to obtain a compact and uniform set for the completion network.

\paragraph{Visual prior encoding.} 
To extract multi-view visual cues, we utilize a PointNet++~\cite{qi2017pointnet++} encoder to extract a comprehensive global point feature $\mathcal{F}_p \in \mathbb{R}^{256}$ from the fused set $\tilde{\mathbf{P}}_{\text{merged}}$. This feature serves as a geometric reference for subsequent multi-view feature aggregation. 

In parallel, we employ an \emph{image pyramid encoder} based on a ResNet backbone~\cite{He_2016_CVPR} to process the $N_v$ completed projections in $\mathcal{V}$. This encoder yields a feature pyramid $\{\mathcal{F}_\text{low} \in \mathbb{R}^{N_v \times 32}, \mathcal{F}_\text{mid} \in \mathbb{R}^{N_v \times 64}, \mathcal{F}_\text{global} \in \mathbb{R}^{N_v \times 128}\}$, capturing semantics ranging from the detailed local geometry and appearance to the coarse global shape.

\begin{figure}[!t]
    \centering
    \begin{overpic}[width=0.95\linewidth]{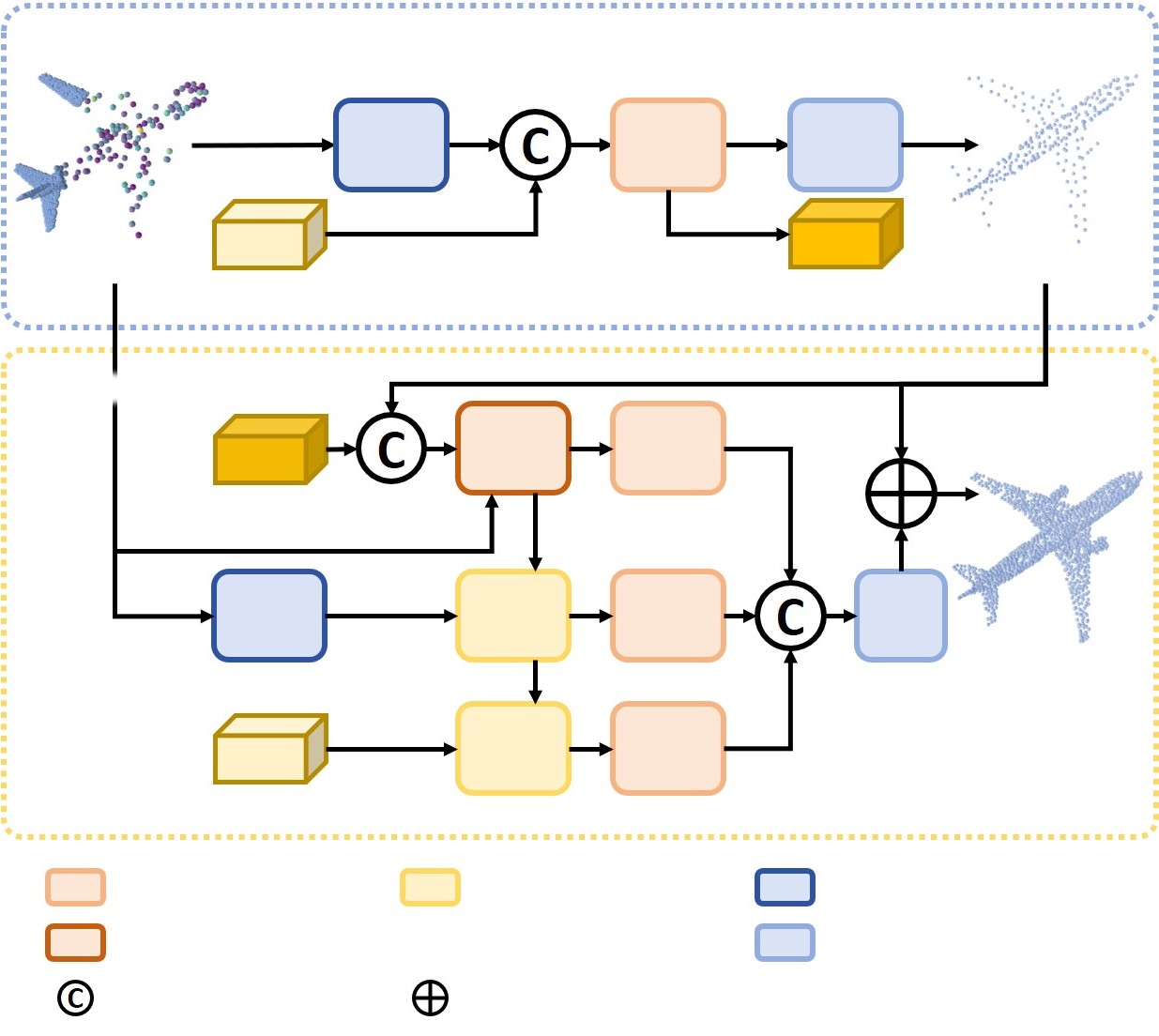} 
    \put(2.5, 84.8){\small \textbf{Coarse Point Generation}}
    \put(2.5, 55.3){\small \textbf{Refinement}}

    \put(4, 66.2){\small $\tilde{\mathbf{P}}_{\text{merged}}$}

    \put(38.7, 78.8) {\small $\mathcal{F}_p$}
    \put(19, 63.5){\small $\mathcal{F}_\text{global}$}
    \put(19, 44.7){\small $\mathcal{F}_\text{fusion}$}
    \put(18, 19){\small $\mathcal{F}_\text{mid}$/$\mathcal{F}_\text{low}$}

    \put(20, 36.6){\small Edge}
    \put(19.5, 33.9){\small Conv}

    \put(30.3, 77.2){\small Point}
    \put(29.6, 74.5){\small Net++}

    \put(33.5, 37.2){\footnotesize Key}
    \put(32.5, 33.4){\footnotesize Value}
    \put(33.5, 25.6){\footnotesize Key}
    \put(32.5, 21.8){\footnotesize Value}

    \put(47, 43){\footnotesize Query}
    \put(47, 30){\footnotesize Query}

    \put(40.8, 49.6){\small IASA}
    \put(42, 35.2){\small CA}
    \put(42, 23.6){\small CA}
    \put(55.6, 75.7){\small SA}
    \put(55.6, 49.6){\small SA}
    \put(55.6, 35.2){\small SA}
    \put(55.6, 23.6){\small SA}

    \put(70.7, 75.7){\small \texttt{MLP}}
    \put(68.8, 63.5){\small $\mathcal{F}_\text{fusion}$}

    \put(70, 41.3){\footnotesize Offset}
    \put(75.3,35.2){\small \texttt{MLP}}

    \put(85, 66.2){\small $\tilde{\mathbf{P}}_\text{coarse}$}
    \put(84, 32){\small $\tilde{\mathbf{P}}_{r1}/\tilde{\mathbf{P}}_{r2}$}

    \put(11, 2.2){\small Concatenation}
    \put(11, 7){\small Incomplete-Aware Self Attention}
    \put(11, 11.8){\small Self Attention}
    \put(41.6, 11.8){\small Cross Attention}
    \put(41.6, 2.2){\small Add}
    \put(72, 11.8){\small Pointcloud Baseline}
    \put(72, 7){\small \texttt{MLP}}
    \end{overpic}
    \vspace{-4mm}
    \caption{\textbf{Coarse points generation and refinement.} We design a coarse point generation module with a cross-attention-based fusion mechanism, and a cascaded two-stage refinement network to produce the final consolidated complete point cloud.}
    \label{images:refinement}
    \vspace{-4mm}
\end{figure}
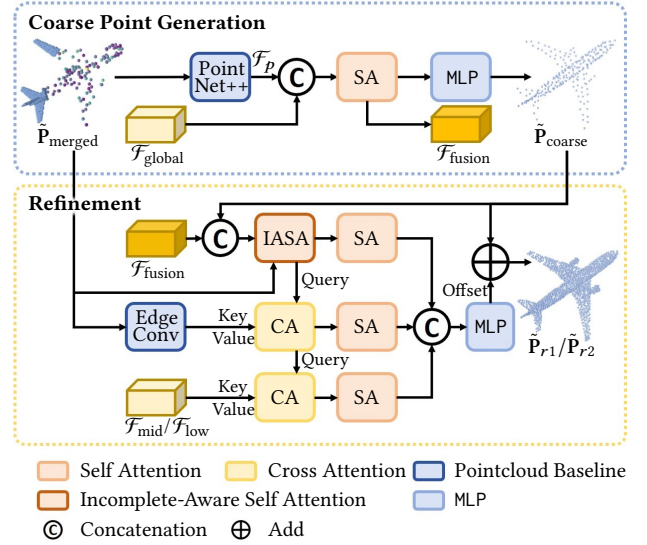

\paragraph{Coarse point generation.} 
Figure~\ref{images:refinement}(top) illustrates the detailed network structure of our point generator.
Specifically, we fuse the high-level multi-view feature $\mathcal{F}_\text{global}$ with the global point feature $\mathcal{F}_p$ via a self-attention (SA) mechanism. 
After fusing, an updated fused feature representation $\mathcal{F}_\text{fusion} \in \mathbb{R}^{256}$ directly predicts an initial coarse completion $\tilde{\mathbf{P}}_\text{coarse} \in \mathbb{R}^{256 \times 3}$ via a few \texttt{MLP} layers. $\tilde{\mathbf{P}}_\text{coarse}$ provides a global scaffold guided by the inferred shape priors, and is structurally plausible, while remaining anchored to the partial points in $\tilde{\mathbf{P}}_{\text{merged}}$.

\paragraph{Cascaded refinement.}
We next design a cascaded refinement network to \emph{upsample} and \emph{consolidate} $\tilde{\mathbf{P}}_\text{coarse}$.
Specifically, as shown in Fig.~\ref{images:refinement}(bottom), the network takes as input $\tilde{\mathbf{P}}_\text{coarse}$, $\tilde{\mathbf{P}}_{\text{merged}}$, $\mathcal{F}_\text{fusion}$, and $\{\mathcal{F}_{low}, \mathcal{F}_{mid}\}$. 
With all the input, we aim to fuse information from both the point clouds and guiding images to drive the completion.
To this end, we design two fusion modes.

\mypar{Mode one: point cloud feature fusion}. Firstly, we concatenate $\mathcal{F}_\text{fusion}$ to the point feature of each point in $\tilde{\mathbf{P}}_\text{coarse}$. Following \cite{zhu2023svdformer}, we incorporate their Incomplete-Aware Self-Attention (IASA) module alongside a standard self-attention mechanism. By explicitly comparing and cross-referencing the coarse prediction $\tilde{\mathbf{P}}_\text{coarse}$ against the merged partial input $\tilde{\mathbf{P}}_\text{merged}$, these modules produce the fused features. With the IASA and SA, we explicitly measure the discrepancy between the confidence-weighted inferred geometry and the coarse shape.
Secondly, we further process $\tilde{\mathbf{P}}_\text{merged}$ using a DGCNN~\cite{wang2019dynamic} feature extractor (\ie EdgeConv) to capture more fine-grained local geometric structures, which is used as key and value for the following cross-attention (CA), where the query tokens are obtained from the fused IASA features. A further SA is employed to refine the second feature. With the CA and SA, an enhanced shape-aware feature is produced.
The two resulting features in this mode are capable of capturing the missing region and the whole shape.

\mypar{Mode two: point and image feature fusion.} In this mode, the pyramid image feature and the point feature go through a cross-attention and self-attention blocks to produce the fused feature. Since the image features capture shape appearance and geometry, especially $\mathcal{F}_\text{low}$, the resulting fused feature is essential to enhance geometric details.

The three features are next concatenated and sent to a few \texttt{MLP} layers to produce upsampled (x8) offset vectors \cite{xiang2021snowflakenet}, which are then added to the eight replicated copies of $\tilde{\mathbf{P}}_\text{coarse}$.
As a result, the first refinement creates an upsampled and consolidated point cloud $\mathbf{P}_{r1} \in \mathbb{R}^{(256\times8) \times 3}$, while the second refinement produces the final resulting point cloud $\mathbf{P}_{r2} \in \mathbb{R}^{(256\times8\times8) \times 3}$. Note that the two refinement blocks have the identical structure, the only difference is that $\tilde{\mathbf{P}}_\text{coarse}$ and $\mathcal{F}_\text{mid}$ are used in the first refinement block, while $\mathbf{P}_{r1}$ and $\mathcal{F}_\text{low}$  are used in the second refinement block.

\subsection{Network Training}
\label{subsec:training}

The completion network, consisting of a point generation block followed by two refinement blocks, is trained end-to-end.
Our objective encourages geometric fidelity to the ground-truth shape while strictly preserving consistency with the observed partial input.
Specifically, we minimize the Chamfer Distance (CD) between the predicted point sets ($\tilde{\mathbf{P}}_\text{coarse}, \tilde{\mathbf{P}}_{r1}, \tilde{\mathbf{P}}_{r2}$) and the ground-truth complete shape $\mathbf{P}_{gt}$. 
The total loss is defined as:
\begin{equation}
    \mathcal{L} = \mathcal{L}_\text{CD}(\tilde{\mathbf{P}}_\text{coarse}, \mathbf{P}_{gt}) + \mathcal{L}_\text{CD}(\tilde{\mathbf{P}}_{r1}, \mathbf{P}_{gt}) + \mathcal{L}_\text{CD}(\tilde{\mathbf{P}}_{r2}, \mathbf{P}_{gt}).
\end{equation}

\section{Experiments and Results}
\label{sec:experiment}

Before delving into the detailed evaluations, it is important to clarify the geometric implications of our partial point cloud generation protocol. Depending on the missing regions, the resulting partial inputs typically fall into two distinct geometric scenarios. The first scenario involves missing regions located primarily in the central or internal parts of the object (\eg, the 25\% and 50\% missing cases in Fig.~\ref{fig:ratio_robustness}). Because the outer extremities remain largely intact, these inputs implicitly preserve the original global scale. The second scenario involves missing regions at the extremities or outer boundaries, resulting in a complete loss of the ground-truth global scale. While our proposed framework is highly robust and seamlessly handles both scenarios, we purposefully highlight the more demanding extremity-missing cases in the majority of our qualitative visualizations (\eg, \cref{fig:pcn_visual,fig:shapenetVisual_visual,fig:generative_completion,fig:noise_robustness,fig:lidar,fig:different_result}). This design choice intuitively illustrates the severe completion challenges introduced by absolute scale ambiguity.

\paragraph{Implementation details.}
Our network is trained for 300 epochs using four NVIDIA RTX 4090 GPUs with a batch size of 24. 
We optimize the parameters using the Adam optimizer with an initial learning rate of $1 \times 10^{-4}$. It is decayed by a factor of 0.7 every 40 epochs on the PCN dataset~\cite{yuan2018pcn} and by a factor of 0.98 every 2 epochs on the ShapeNet-55 dataset~\cite{chang2015shapenet}.
For all experiments, the hyperparameters are fixed at $\tau_{\text{overlap}}=0.7$, $\tau_{\text{prob}}=0.8$, $\Delta=0.02$, $\delta_{\text{diff}}=30$, $N=2048$. The number of rendering viewpoints is set to $13$, and the render image size is $256\times256$.

\paragraph{Dataset}
To train and test the network, we employ two benchmark datasets.
The PCN dataset is a subset of the ShapeNet dataset, comprising eight object categories where each ground-truth point cloud consists of 16,384 points. It contains a total of 30,974 samples, which are partitioned into training, validation, and test sets following the exact same split ratio as PCDreamer~\cite{wei2025pcdreamer}.
We construct the dataset as follows. Given a complete shape, we randomly select a seed point and remove a specific proportion of its nearest neighbors using the $K$-nearest neighbor (KNN) algorithm. By varying the removal ratios (25\%, 50\%, and 75\% of the total points), we generate partial point clouds with different levels of structural incompleteness. 
At inference time, the VLM inpaints the missing region roughly.
To simulate the variation at inference time, in the data construction, we apply a similar process as prior acquisition shown in Fig.~\ref{images:pipeline}(top). 
In the end, each data sample contains the multi-view images, the merged point cloud, the partial points, and the GT points for training the completion network.
Please refer to the \emph{supplementary} for more construction details. 
In contrast to the PCN dataset, which is restricted to only eight categories, ShapeNet-55 provides a significantly larger and more diverse repository encompassing 55 distinct object classes. To evaluate the scalability and robustness of our proposed model on this large-scale benchmark, we strictly follow the same data construction protocol as aforementioned.

\paragraph{Evaluation Metric}
To quantitatively evaluate the completion performance, we employ three standard metrics: Chamfer Distance (CD), F-Score, and Density-aware Chamfer Distance (DCD) \cite{wu2021balanced}. CD measures the global geometric distance between predicted and ground-truth point sets, but is sensitive to outliers and ignores local density variations.
F-Score supplements CD to comprehensively assess visual fidelity and surface reconstruction quality.
Unlike vanilla CD, DCD is designed to be more sensitive to the spatial distribution, penalizing uneven point clusters and encouraging more realistic geometric detail.

\subsection{Comparison: Direct Point Cloud Completion}

To verify the effectiveness of our proposed method, we conduct a comprehensive comparison with several state-of-the-art completion frameworks, including AdaPoinTr \cite{10232862}, SVDFormer \cite{zhu2023svdformer}, CRA-PCN \cite{rong2024cra}, GeoFormer \cite{yu2024geoformer}, SIMECO \cite{wang2025learning}, and PCDreamer \cite{wei2025pcdreamer}. 
All baseline networks are retrained from scratch on our benchmark using their default settings. We report results under two preprocessing protocols: 
\begin{itemize}
    \item[(1)] \emph{GT-scale-free normalization (unknown scale).} We normalize each partial input using the scale inferred by our \emph{prior acquisition} module, and apply the \emph{same} estimated scale to all competing methods. 
    Note that SIMECO \cite{wang2025learning} addresses the GT scale ambiguity, we evaluate them exclusively under the \emph{GT-scale-free} setting (without normalizing their partial inputs using the scale inferred by our \emph{prior acquisition} module).
    Unless otherwise specified, the visual results shown in this paper are produced under this setting.

    \item[(2)] \emph{GT-scale-tied normalization (oracle scale).} We additionally report results under the standard synthetic protocol, where normalization uses the ground-truth full-shape scale. Note that, although the setting is impractical, it provides a performance upper bound for all methods, and also \emph{validates the effectiveness of our completion network}.
\end{itemize}

We additionally consider a third setting, termed \emph{on-the-fly partial-based normalization}, where the input partial observation is normalized using only its own spatial extent at inference time.
Under this protocol, existing baselines---originally trained with GT-scale-tied normalization and thus relying on oracle scale cues---suffer from a substantial distribution shift and yield severely degraded completions.
For completeness, we report preliminary qualitative results under this setting in the supplementary; however, due to the consistently poor performance and the lack of meaningful quantitative comparability, we do not include this setting in the main comparison.

\begin{figure*}[h]
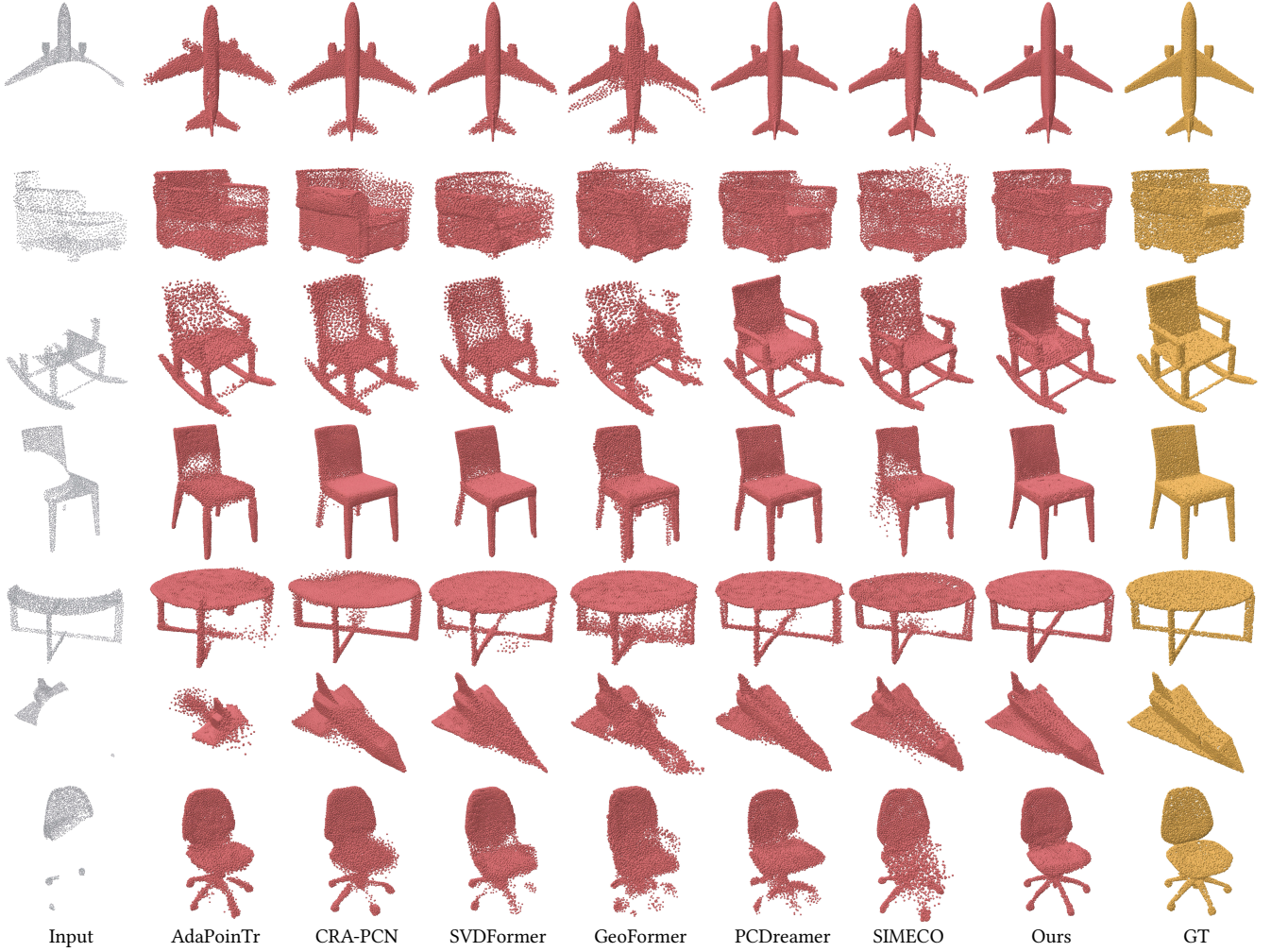

    \centering
    \begin{overpic}[width=0.99\textwidth]{images/pcn_result_wo_words.jpg} 
    \put(4.2, -1.5){\small Input}
    \put(13.8, -1.5){\small AdaPoinTr}
    \put(25.2, -1.5){\small CRA-PCN}
    \put(35.8, -1.5){\small SVDFormer}
    \put(47.2, -1.5){\small GeoFormer}
    \put(58.3, -1.5){\small PCDreamer}
    \put(69.2, -1.5){\small SIMECO}
    \put(81.7, -1.5){\small Ours}
    \put(93.7, -1.5){\small GT}
    \end{overpic}
    \caption{\textbf{Qualitative comparison on the PCN dataset under the GT-scale-free setting}. Given the inputs with varying missing regions, baseline methods produce inferior completion results with clear artifacts, while our method completes objects faithfully.}
  \label{fig:pcn_visual}
\end{figure*}

\begin{table*}[!t]
    \centering
    \caption{\textbf{Quantitative comparison on the PCN dataset under the \emph{GT-scale-free} and \emph{GT-scale-tied} settings}. Shapes from eight categories are used in the evaluation. We report the CD-Avg$\downarrow$ for each category, and additionally report the CD-Avg$\downarrow$, DCD-Avg$\downarrow$, F1$\uparrow$ across all categories. For each cell, the two numbers are for the GT-scale-free and GT-scale-tied settings, respectively.
    We use the \textcolor{bestbg}{dark green} and \textcolor{bestblue}{dark blue} to indicate the best results for evaluations under GT-scale-free and GT-scale-tied settings, while using the \textcolor{secondbg}{light green} and \textcolor{secondblue}{light blue} to indicate the second best. 
    }
    \label{tab:pcn_results}
    \vspace{-2mm}
    \resizebox{\textwidth}{!}{
    \begin{tabular}{l|cccccccc|ccc}
\toprule
        Methods & Plane & Cabinet & Car & Chair & Lamp & Couch & Table & Boat & CD-Avg$\downarrow$ & DCD-Avg$\downarrow$ & F1$\uparrow$ \\
        \midrule
        AdaPoinTr* & 4.28 / 3.81 & 8.11 / 7.75 & 7.58 / 7.12 & \secondboxg{6.99} / 6.97 & 9.45 / 7.96 & \secondboxg{6.51} / 6.32 & 6.71 / 5.98 & 6.94 / 5.75 & 7.07 / 6.46 & 0.530 / 0.528 & 0.826 / 0.852 \\
        
        SVDFormer* & 4.38 / 3.95 & 7.98 / 7.39 & 7.69 / 6.98 & 7.13 / 6.97 & 9.42 / 8.14 & 7.06 / 6.21 & 6.77 / 5.73 & 6.96 / 5.82 & 7.18 / 6.40 & 0.532 / 0.510 & 0.818 / 0.857 \\
        
        CRA-PCN* & \secondboxg{4.20} / 4.08 & \secondboxg{7.55} / \bestboxb{6.84} & 7.65 / 7.15 & 7.08 / \secondboxb{6.13} & \secondboxg{8.89} / \secondboxb{7.12} & 6.67 / 6.04 & \secondboxg{6.62} / \secondboxb{5.50} & \secondboxg{6.83} / 5.70 & \secondboxg{6.94} / 6.10 & \secondboxg{0.523} / 0.507 & \secondboxg{0.834} / 0.862 \\
        
        GeoFormer* & 4.87 / 3.90 & 7.85 / 7.18 & 7.77 / 6.90 & 7.50 / 6.98 & 9.17 / 8.09 & 6.76 / \secondboxb{5.93} & 6.98 / 5.56 & 7.09 / 5.78 & 7.25 / 6.29 & 0.535 / 0.508 & 0.814 / 0.869 \\
        
        PCDreamer* & 4.23 / \secondboxb{3.75} & 7.88 / \secondboxb{6.93} & \secondboxg{7.53} / \secondboxb{6.88} & 7.45 / 6.14 & 9.03 / 7.81 & 6.66 / 5.94 & 6.76 / 5.52 & 6.85 / \secondboxb{5.69} & 7.05 / \secondboxb{6.08} & 0.528 / \secondboxb{0.503} & 0.829 / \secondboxb{0.871} \\
        
        SIMECO & 4.72 / - - - & 8.43 / - - - & 8.03 / - - - & 6.97 / - - - & 10.26 / - - - & 7.45 / - - - & 7.71 / - - - & 6.97 / - - - & 7.56 / - - - & - - - / - - - & 0.774 / - - - \\
        
        \midrule
        \textbf{Ours} & \bestboxg{3.88} / \bestboxb{3.56} & \bestboxg{7.39} / 7.03 & \bestboxg{7.08} / \bestboxb{6.84} & \bestboxg{6.49} / \bestboxb{6.09} & \bestboxg{7.34} / \bestboxb{7.10} & \bestboxg{6.18} / \bestboxb{5.87} & \bestboxg{5.72} / \bestboxb{5.44} & \bestboxg{5.89} / \bestboxb{5.66} & \bestboxg{6.21} / \bestboxb{5.95} & \bestboxg{0.514} / \bestboxb{0.499} & \bestboxg{0.871} / \bestboxb{0.883} \\ 
        \bottomrule
    \end{tabular}
    }
\end{table*}

\paragraph{Evaluation on the PCN dataset.}
Table~\ref{tab:pcn_results} reports results under both the GT-scale-free and the conventional GT-scale-tied settings. \mname delivers the strongest performance in both protocols and is the only method that consistently ranks first across all eight categories under the GT-scale-free setting.
This setting is practical for real-world applications, and our superior performance indicates that our completion network does not rely on GT-scale-scale canonicalization to recover global extent, structure, and geometry. 
The gains are most evident from categories that are sensitive to scale cues (e.g., \emph{Plane}, \emph{Lamp}, and \emph{Boat}), where large missing regions often disrupt the object span and amplify the ambiguity of completion. 
More importantly, \mname is substantially more stable when moving from GT-scale-tied to GT-scale-free preprocessing. 
While prior methods exhibit a clear performance drop once the oracle scale cue is removed, \mname maintains competitive accuracy with only a modest degradation.

Figure~\ref{fig:pcn_visual} shows typical examples from all competing methods. 
The first five rows correspond to partial inputs with 50\% missing, while the last two rows use the more challenging 75\% setting. 
When the partial input still preserves a reliable scale cue (\eg the \emph{Chair} in the fourth row, where the missing region is mostly internal), most methods perform reasonably well. In contrast, at 75\% missing, large holes often remove key evidence about the global extent (\eg the \emph{Plane} in the sixth row). In this case, many baselines degrade and produce noisy points, over-completion, or missing parts.

\begin{table*}[t]
    \centering
    \caption{\textbf{Quantitative comparison on the ShapeNet-55 dataset under the \emph{GT-scale-free} and \emph{GT-scale-tied} settings}. Shapes from ten categories are used in the evaluation. We report the CD-Avg$\downarrow$ for each category, and additionally report the CD-Avg$\downarrow$, DCD-Avg$\downarrow$, F1$\uparrow$ across all categories. For each cell, the two numbers are for the GT-scale-free and GT-scale-tied settings, respectively.
    We use the \textcolor{bestbg}{dark green} and \textcolor{bestblue}{dark blue} to indicate the best results for evaluations under GT-scale-free and GT-scale-tied settings, while using the \textcolor{secondbg}{light green} and \textcolor{secondblue}{light blue} to indicate the second best.
    }
    \label{tab:shapenet55_results}
    \vspace{-2mm}
    \resizebox{1.0\textwidth}{!}{
    \begin{tabular}{l|ccccc|ccccc|ccc}
        \toprule
        Methods & Table & Chair & Plane & Car & Sofa & Bird house & Bag & Microwave & Mailbox & Dishwasher & CD-Avg$\downarrow$ & DCD-Avg$\downarrow$ & F1$\uparrow$ \\
        \midrule
        AdaPointr* & 1.28 / 0.84 & \secondboxg{0.88} / 0.81 & \secondboxg{0.55} / 0.47 & \secondboxg{0.96} / 0.93 & 1.17 / 0.86 & 1.62 / 1.34 & 1.13 / 0.74 & 2.08 / 1.41 & \secondboxg{0.85} / 0.71 & 1.51 / 1.15 & 1.40 / 1.08 & 0.658 / 0.619 & 0.298 / 0.408 \\
        SVDFormer* & 1.38 / 0.78 & 1.05 / 0.77 & 0.64 / 0.45 & 1.05 / \secondboxb{0.75} & 1.27 / 0.86 & 1.87 / 1.19 & 1.51 / 0.72 & 2.21 / 1.19 & 1.14 / 0.59 & 1.77 / 1.04 & 1.45 / 0.93 & 0.661 / 0.568 & 0.288 / 0.437 \\
        CRA-PCN* & 1.54 / 0.83 & 1.07 / 0.83 & 0.62 / 0.58 & 1.09 / 0.96 & 1.24 / 0.94 & 1.78 / 1.38 & 1.21 / 0.78 & 2.29 / 1.38 & 1.18 / 0.63 & 1.84 / 1.59 & 1.37 / 1.10 & 0.657 / 0.623 & 0.295 / 0.405 \\
        GeoFormer* & 1.42 / 0.75 & 0.99 / \secondboxb{0.76} & 0.69 / 0.42 & 1.01 / 0.84 & 1.18 / \secondboxb{0.83} & 1.54 / \secondboxb{1.12} & 1.12 / 0.68 & \secondboxg{1.97} / 1.14 & 1.18 / 0.63 & \secondboxg{1.45} / 0.98 & 1.31 / 0.88 & \secondboxg{0.624} / 0.556 & \secondboxg{0.335} / \secondboxb{0.454} \\
        PCDreamer* & \secondboxg{1.12}/ \secondboxb{0.72} & 0.98 / \bestboxb{0.74} & 0.59 / \secondboxb{0.41} & 0.99 / 0.79 & \secondboxg{1.13} / 0.85 & \secondboxg{1.46} / 1.26 & \secondboxg{1.03} / \secondboxb{0.65} & 2.17 / \secondboxb{1.09} & 0.91 / \secondboxb{0.58} & 1.54 / \secondboxb{0.96} & \secondboxg{1.29} / \secondboxb{0.86} & 0.628 / \secondboxb{0.551} & 0.322 / 0.442 \\
        \midrule
        \textbf{Ours} & \bestboxg{0.89} / \bestboxb{0.69} & \bestboxg{0.81} / \bestboxb{0.74} & \bestboxg{0.51} / \bestboxb{0.40} & \bestboxg{0.83} / \bestboxb{0.73} & \bestboxg{0.98} / \bestboxb{0.78} & \bestboxg{1.17} / \bestboxb{1.05} & \bestboxg{0.90} / \bestboxb{0.58} & \bestboxg{1.41} / \bestboxb{1.02} & \bestboxg{0.64} / \bestboxb{0.53} & \bestboxg{1.22} / \bestboxb{0.92} & \bestboxg{1.18} / \bestboxb{0.84} & \bestboxg{0.615} / \bestboxb{0.549} & \bestboxg{0.347} / \bestboxb{0.458} \\
        \bottomrule
    \end{tabular}
    }
\end{table*}

\begin{figure*}[!t]
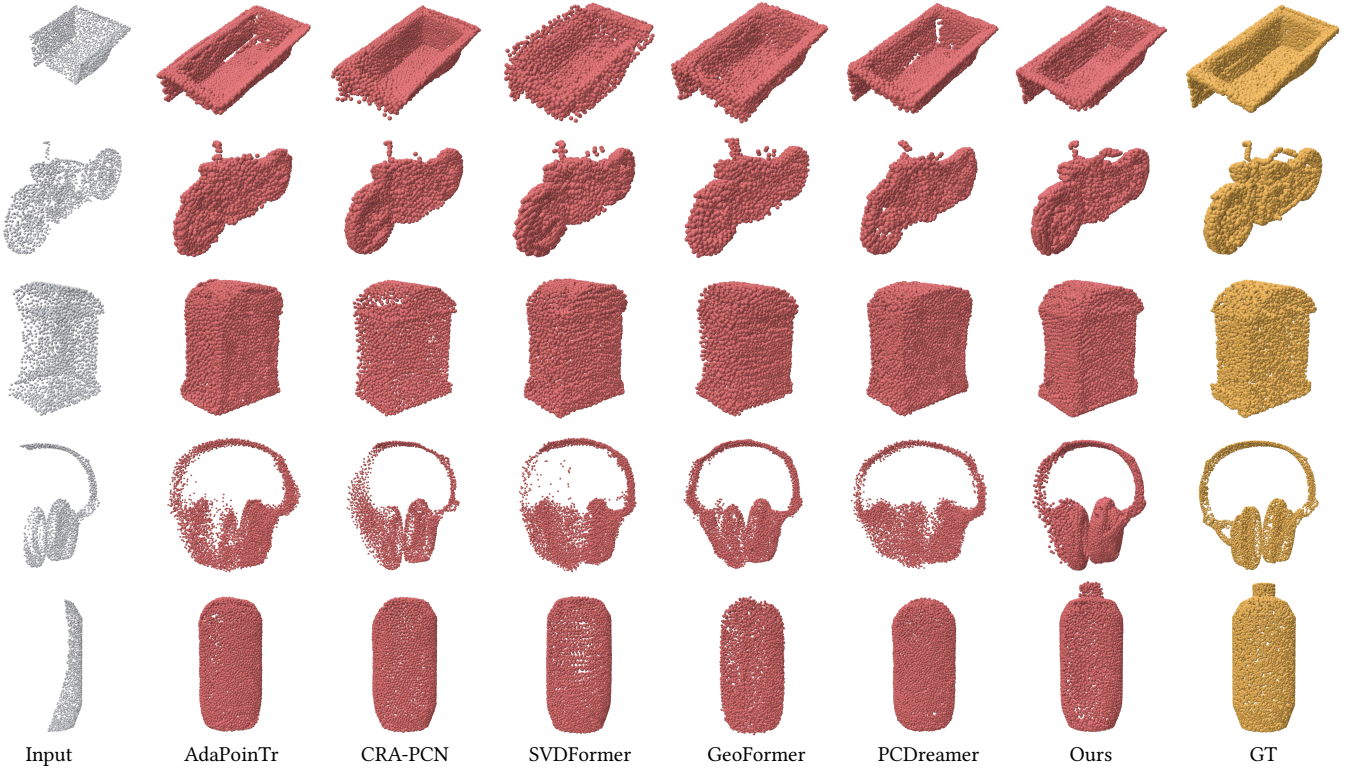

    \centering
    \begin{overpic}[width=0.99\textwidth]{images/shapenet_result_wo_words} 
    \put(2, -1.5){\small Input}
    \put(13.8, -1.5){\small AdaPoinTr}
    \put(27, -1.5){\small CRA-PCN}
    \put(39.5, -1.5){\small SVDFormer}
    \put(52.8, -1.5){\small GeoFormer}
    \put(65.5, -1.5){\small PCDreamer}
    \put(79.8, -1.5){\small Ours}
    \put(93.2, -1.5){\small GT}
    \end{overpic}
    \caption{\textbf{Qualitative comparison on the ShapeNet-55 dataset under the GT-scale-free setting}. Given the inputs with varying missing regions, baseline methods produce inferior completion results with clear artifacts, while our method completes objects faithfully.}
  \label{fig:shapenetVisual_visual}
\end{figure*}

\paragraph{Evaluation on the ShapeNet-55 dataset.}
Other than the PCN dataset, we have evaluated the performance of all the methods on the ShapeNet-55 dataset. 
Specifically, Tab~\ref{tab:shapenet55_results} reports results under both the GT-scale-free and GT-scale-tied settings. Compared to PCN, ShapeNet-55 spans more categories and exhibits larger intra-class shape variation, which makes GT-scale-free completion more challenging. 
Despite this broader diversity, \mname still achieves the strongest overall results in both protocols and remains stable when moving from GT-scale-tied to GT-scale-free preprocessing.

Visual results are shown in Fig.~\ref{fig:shapenetVisual_visual}. Our method produces superior completions.

\subsection{Comparison: Generative Point Cloud Completion}
We further compare \mname with recent image-to-3D generators, including Hi3DGen~\cite{Ye_2025_ICCV} and Trellis~\cite{Xiang_2025_CVPR}. 
Both models are pretrained to synthesize complete 3D assets from a single RGB input, thus providing a strong alternative solution for completion. 
For a fair comparison, we feed the 13 VLM-completed images produced by our pipeline into these models to generate meshes, choose the best result from the 13 candidates, and uniformly sample points from the output via Poisson disk sampling for evaluation. Note that we run several rounds of each generation, since they are sometimes unstable to produce reasonable results.

\begin{figure}[!h]
    \centering
    \begin{overpic}[width=0.99\linewidth]{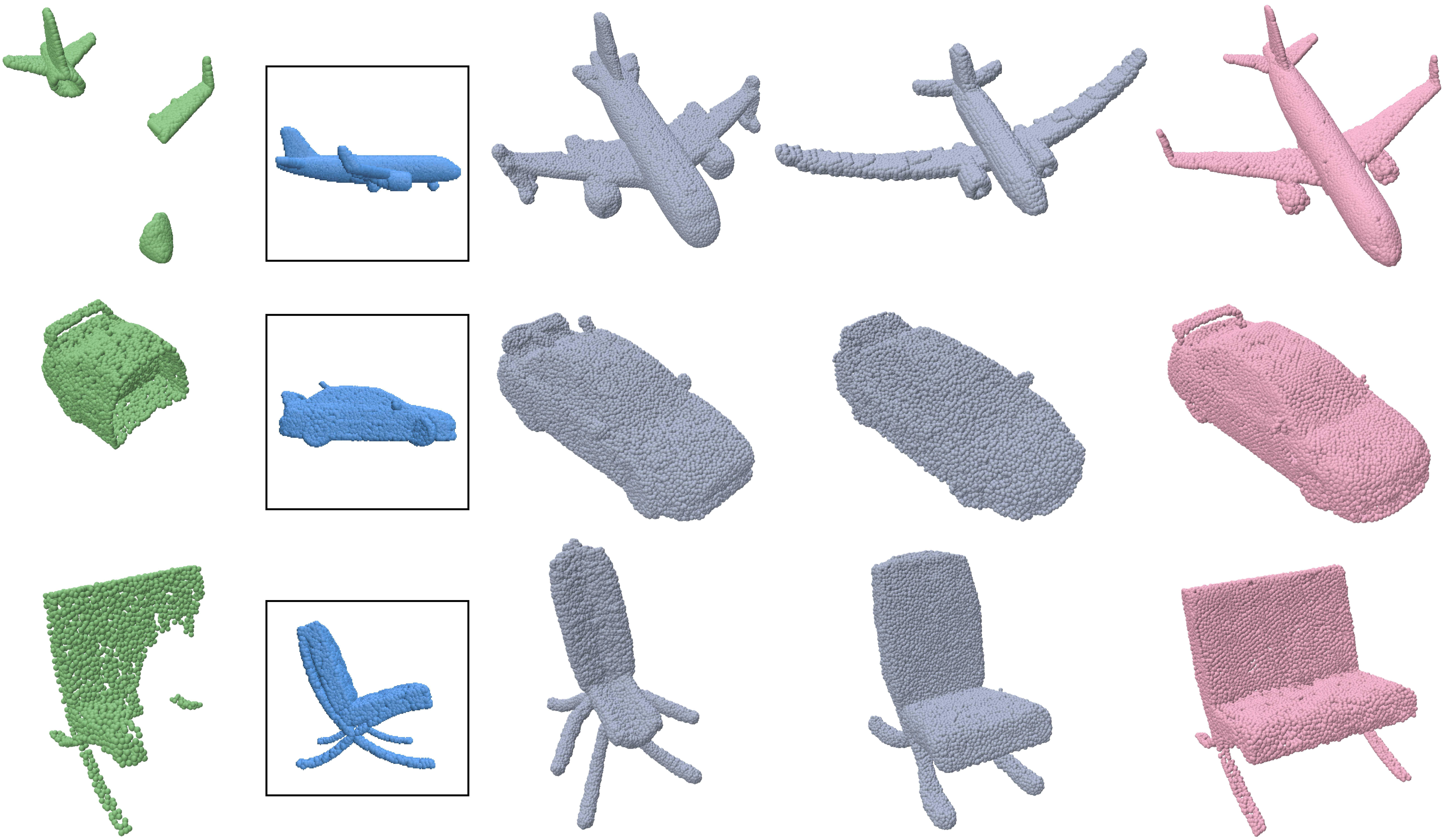} 
    \put(6, -3){\small Input}
    \put(21, -3){\small Image}
    \put(37, -3){\small Hi3DGen}
    \put(62.4, -3){\small Trellis}
    \put(86.9, -3){\small Ours}
    \end{overpic}
    \caption{\textbf{Generative completion.} 
    Given the partial input, we use VLMs to complete partial renderings and send them to two representative image-to-3D generation models to reconstruct the full shape. Although their results are structurally meaningful and visually plausible, they fail to align with the partial observation.
    Our results are of high fidelity and respect the inputs. 
    }
    \label{fig:generative_completion}
\end{figure}

Visual comparison is shown in Fig.~\ref{fig:generative_completion}, where these generative baselines often yield plausible global shapes but misaligned geometry compared with the partial observation, such as the wings of the airplane or thin hollow structures on car spoilers, due to the limited information in a single view. In the future, a more powerful generative model taking as input multi-view images could potentially solve this issue.
In contrast, \mname leverages consistent evidence across multiple views and recovers more faithful shapes with higher fidelity.

\begin{table}[!t]
    \centering
    \caption{\textbf{Statistical ablation results}. Statistics on CD-Avg$\downarrow$, DCD-Avg$\downarrow$, F1$\uparrow$ are reported from the ablated variants of our method.}
    \label{tab:ablation_components}
    \vspace{-2mm}
    \resizebox{0.8\linewidth}{!}{
    \begin{tabular}{cc|ccc}
        \toprule
        Conf. Att. & Hier. Feat. & CD-Avg$\downarrow$ & DCD-Avg$\downarrow$ & F1$\uparrow$ \\
        \midrule
        \rowcolor{gray!10} & \checkmark & 6.32 & 0.519 & 0.866 \\ 
        \checkmark &  & 6.41 & 0.522 & 0.863 \\ 
        \rowcolor{gray!10} \checkmark & \checkmark & \textbf{6.21} & \textbf{0.514} & \textbf{0.871} \\ 
        \bottomrule
    \end{tabular}
    }
    \vspace{-3mm}
\end{table}

\subsection{Ablation Study}
\label{sec:ablation}
We conduct extensive ablation experiments to validate our core design choices. All quantitative and qualitative evaluations are performed on the PCN dataset.

\begin{figure}[!t]
    \centering
    \begin{overpic}[width=0.9\linewidth]{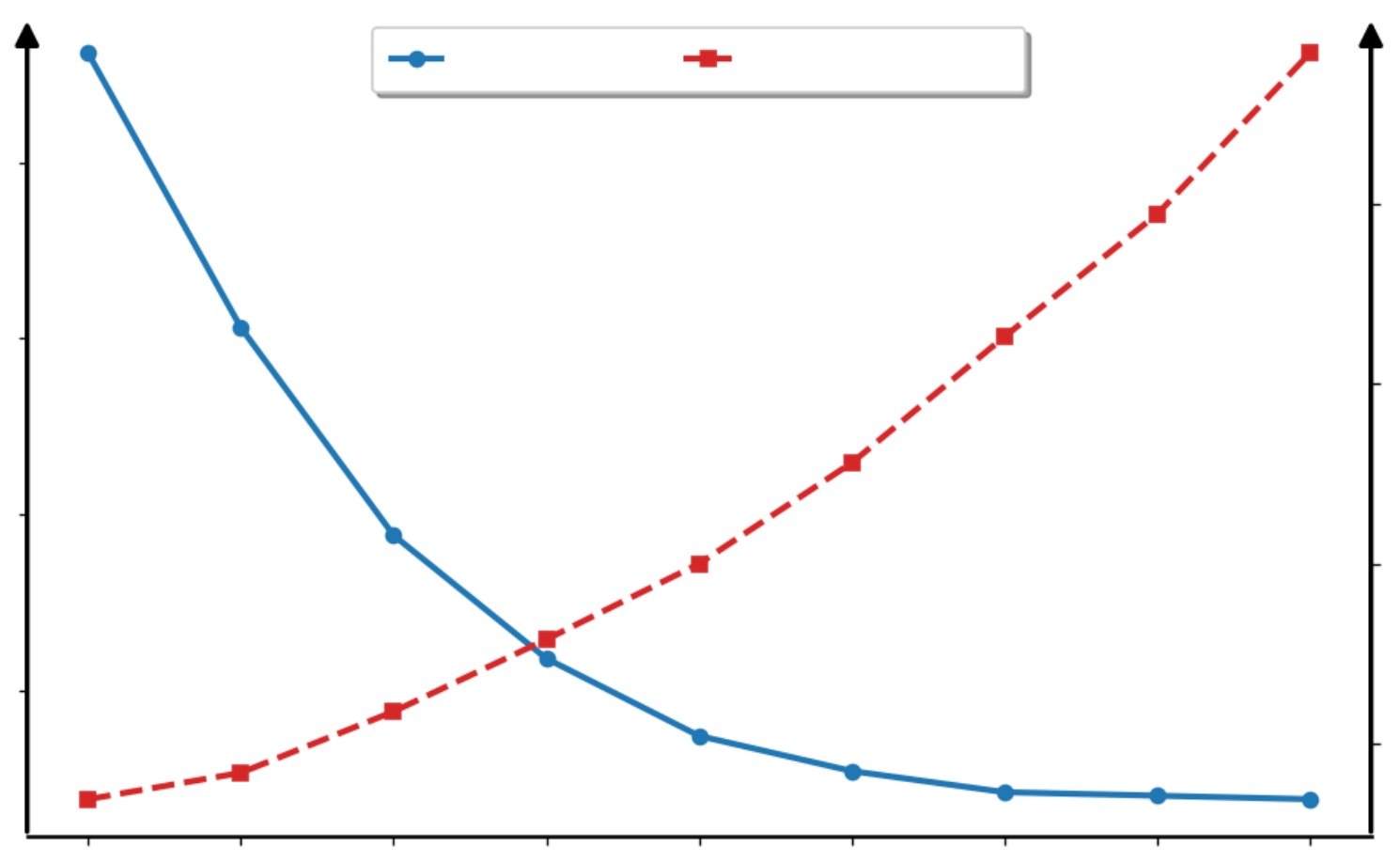}
    \put(33.5, 56.2){\scriptsize CD Value}
    \put(54, 56.2){\scriptsize Inference Time}
    \put(5.8, -1){\scriptsize 1}
    \put(16.8, -1){\scriptsize 3}
    \put(27.7, -1){\scriptsize 5}
    \put(38.6, -1){\scriptsize 7}
    \put(49.6, -1){\scriptsize 9}
    \put(59.7, -1){\scriptsize 11}
    \put(70.8, -1){\scriptsize 13}
    \put(81.8, -1){\scriptsize 15}
    \put(92.5, -1){\scriptsize 17}
    \put(-2, 11.5){\scriptsize 6.5}
    \put(-2, 24.3){\scriptsize 7.0}
    \put(-2, 36.5){\scriptsize 7.5}
    \put(-2, 49.1){\scriptsize 8.0}
    \put(99.3, 7.6){\scriptsize 0.3}
    \put(99.3, 20.5){\scriptsize 0.45}
    \put(99.3, 33.3){\scriptsize 0.6}
    \put(99.3, 46.1){\scriptsize 0.75}
    \put(42, -4.6){\scriptsize Number of views}
    \put(-5, 30){\makebox(0,0){\rotatebox{90}{\scriptsize CD Value}}}
    \put(106.5, 30){\makebox(0,0){\rotatebox{90}{\scriptsize Inference Time (s)}}}
    \end{overpic}
    \vspace{1mm}
    \caption{\textbf{Viewpoint numbers vs. CD and inference time.}
    When increasing the number of viewpoints, on the one hand, the completion quality improves sharply at first and then gradually flattens out. On the other hand, the inference time of VLMs increases significantly.
    }
    \label{fig:number_of_views}
\end{figure}

\paragraph{Impact of the number of viewpoints.}
Image-based guidance is critical for high-fidelity point cloud completion, and the number of views is a key hyperparameter affecting the final results. To investigate this, we generated 1-18 candidate viewpoints for each object and systematically evaluated model performance across different view counts. Fig.~\ref{fig:number_of_views} illustrates the relationship between the number of views and the inference time. We observed that the error decreases significantly as more views are added, with $|\mathcal{V}|=13$ providing the optimal balance between geometric accuracy and computational efficiency.

\paragraph{Efficacy of the confidence attribute.}
The confidence attribute (denoted as Conf. Att.) is introduced to modulate the contribution of prior candidates, down-weighting uncertain regions while preserving the reliable evidence and partial geometry. We ablate this component by removing the confidence values and feeding the network only the merged point set with uniform weights. As shown in Tab.~\ref{tab:ablation_components}, performance drops consistently when the confidence attribute is disabled, indicating that explicitly encoding confidence is important for suppressing noisy prior candidates and improving the stability of prior-guided completion under GT-scale-free preprocessing.

\paragraph{Impact of hierarchical image features.}
To assess the role of the hierarchical image features (denoted as Hier. Feat.), we ablate our feature injection by using only the global image feature ($\mathcal{F}_\text{global}$) in the refinement block, while keeping all other settings unchanged. As reported in Tab.~\ref{tab:ablation_components}, this variant consistently underperforms the full model, confirming that hierarchical features are important for detailed reconstruction. In particular, global features primarily constrain the overall extent and coarse structure, whereas mid/low-level cues provide the local geometric evidence needed to refine thin parts and recover fine-grained details.

\subsection{Discussion and Application}

\begin{table*}[!t]
    \centering
    \caption{\textbf{Quantitative comparison using single-view point clouds rendered from depth maps.}
    Following the same data generation protocol as PCDreamer~\cite{wei2025pcdreamer}, the partial inputs are acquired from single-view depth renderings. We report the CD-Avg$\downarrow$ for each category, and additionally report the CD-Avg$\downarrow$, DCD-Avg$\downarrow$, and F1$\uparrow$ across all categories. For each cell, the two numbers are for the GT-scale-free and GT-scale-tied settings, respectively.
    We use \textcolor{bestbg}{dark green} and \textcolor{bestblue}{dark blue} to indicate the best results for evaluations under GT-scale-free and GT-scale-tied settings, while using \textcolor{secondbg}{light green} and \textcolor{secondblue}{light blue} to indicate the second best. 
    }
    \label{tab:single_view_results}
    \vspace{-2mm}
    \resizebox{\textwidth}{!}{
    \begin{tabular}{l|cccccccc|ccc}
        \toprule
        Methods & Plane & Cabinet & Car & Chair & Lamp & Couch & Table & Boat & CD-Avg$\downarrow$ & DCD-Avg$\downarrow$ & F1$\uparrow$ \\
        \midrule
        AdaPoinTr* & 4.86 / 3.53 & 9.34 / 8.99 & 7.37 / 6.92 & \secondboxg{7.84} / \bestboxb{6.57} & 9.50 / \secondboxb{5.57} & \secondboxg{9.68} / 8.34 & 7.79 / \bestboxb{5.96} & 7.12 / 5.93 & \secondboxg{7.94} / \secondboxb{6.48} & \secondboxg{0.595} / 0.530 & \secondboxg{0.693} / 0.852 \\
        
        SVDFormer* & 4.63 / 3.68 & 9.27 / 8.73 & 7.52 / 7.10 & 8.64 / 6.95 & 9.63 / 5.64 & 9.94 / 8.58 & \secondboxg{7.68} / 6.26 & 7.83 / 5.91 & 8.14 / 6.61 & 0.673 / 0.534 & 0.674 / 0.848 \\
        
        CRA-PCN* & 4.82 / 3.62 & \secondboxg{9.15} / 8.77 & 7.30 / 7.00 & 8.74 / 6.92 & 9.83 / \bestboxb{5.46} & 10.53 / 8.59 & 8.05 / 6.27 & 8.23 / \secondboxb{5.86} & 8.33 / 6.56 & 0.631 / 0.537 & 0.672 / 0.846 \\
        
        GeoFormer* & 5.16 / 3.70 & 10.03 / 9.08 & 7.24 / 6.93 & 9.92 / 7.19 & 10.30 / 6.30 & 10.67 / 8.97 & 8.61 / 7.14 & 8.13 / 5.87 & 8.75 / 6.90 & 0.697 / 0.563 & 0.659 / 0.821 \\
        
        PCDreamer* & \secondboxg{4.26} / \secondboxb{3.52} & 9.25 / \secondboxb{8.72} & 7.21 / \bestboxb{6.89} & 7.96 / 6.71 & 9.81 / 5.64 & 9.69 / \secondboxb{8.32} & 7.94 / 6.24 & 7.62 / \bestboxb{5.84} & 7.97 / 6.49 & 0.614 / \secondboxb{0.518} & 0.681 / \bestboxb{0.859} \\
        
        SIMECO & 4.80 / - - - & 12.11 / - - - & \bestboxg{7.13} / - - - & 8.88 / - - - & \bestboxg{7.62} / - - - & 11.32 / - - - & 8.07 / - - - & \secondboxg{6.50} / - - - & 8.30 / - - - & - - - / - - - & 0.674 / - - - \\
        
        \midrule
        \textbf{Ours} & \bestboxg{4.10} / \bestboxb{3.46} & \bestboxg{8.43} / \bestboxb{8.31} & \secondboxg{7.16} / \secondboxb{6.90} & \bestboxg{6.97} / \secondboxb{6.69} & \secondboxg{8.21} / 5.85 & \bestboxg{8.87} / \bestboxb{7.93} & \bestboxg{7.27} / \secondboxb{6.21} & \bestboxg{6.48} / 5.93 & \bestboxg{7.19} / \bestboxb{6.41} & \bestboxg{0.538} / \bestboxb{0.513} & \bestboxg{0.722} / \secondboxb{0.855} \\
        \bottomrule
    \end{tabular}
    }
\end{table*}

\paragraph{Evaluation on single-view scanned partial points}
In addition to generating partial point clouds via point cropping, we further evaluate our method using single-view point clouds from the virtual scanning processing, strictly following the data generation protocol of PCDreamer~\cite{wei2025pcdreamer}. 
To comprehensively validate the effectiveness of our approach, this evaluation is conducted under both the traditional \emph{GT-scale-tied} setting and the more challenging \emph{GT-scale-free} setting. As shown in Tab.~\ref{tab:single_view_results}, under the GT-scale-tied assumption, our method consistently outperforms other state-of-the-art completion networks. This demonstrates that our core architecture possesses robust feature extraction and geometric reasoning capabilities, even when confronted with highly self-occluded single-view observations. Furthermore, when extended to the GT-scale-free setting, our approach maintains its overall superiority. Although it yields slightly sub-optimal results in a small minority of specific object categories, our method secures the best overall performance across all average metrics (i.e., CD-Avg, DCD-Avg, and F1). This confirms that our framework can effectively leverage visual priors to overcome absolute scale ambiguity, achieving robust and comprehensive completions across diverse shapes.

\begin{figure}[!t]
    \centering
    \begin{overpic}[width=0.99\linewidth]{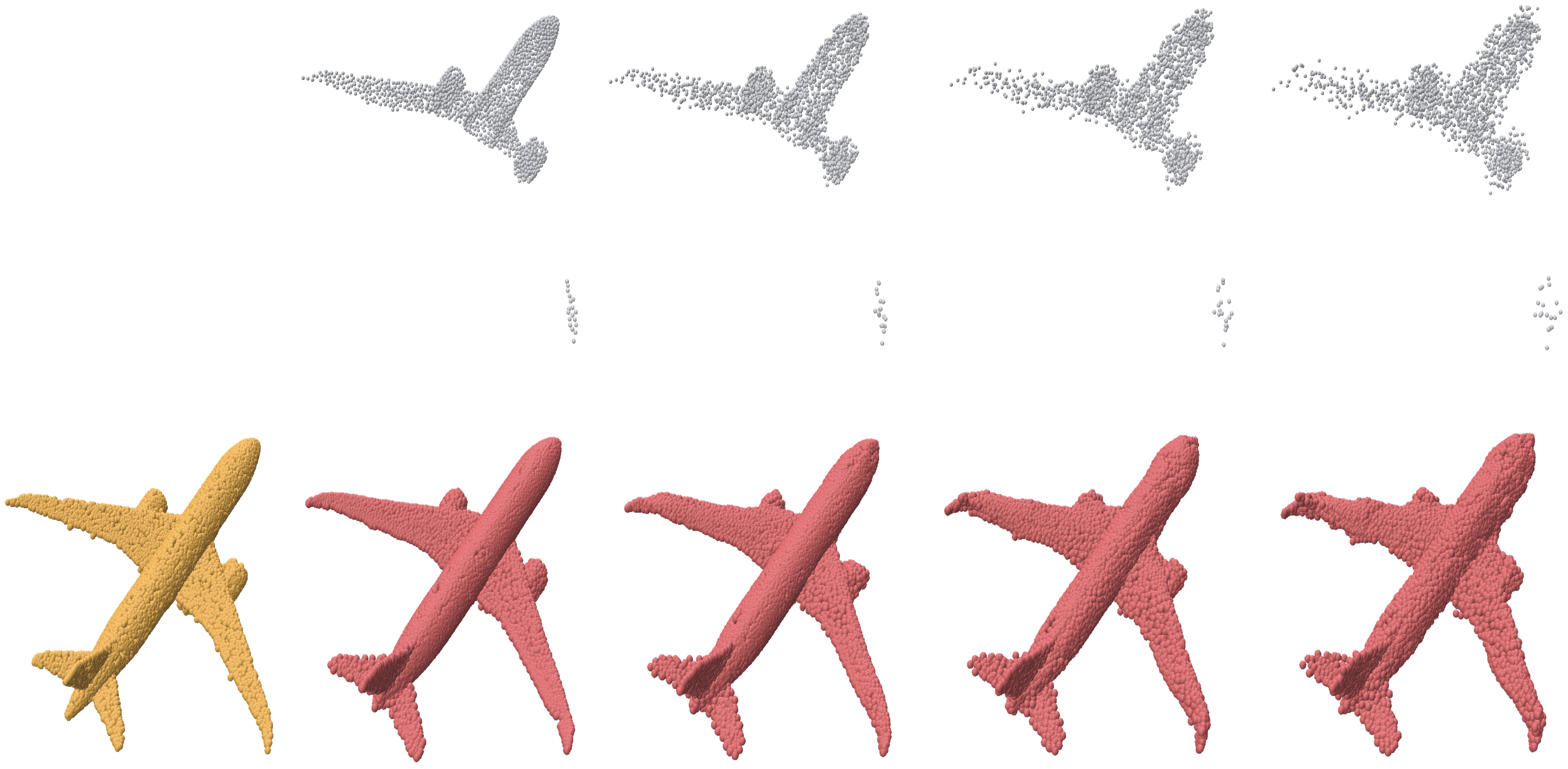}
    \put(7, -3){\small GT}
    \put(23, -3){\small 0\% noise}
    \put(41.5, -3){\small 0.5\% noise}
    \put(61.5, -3){\small 1.0\% noise}
    \put(83.3, -3){\small 1.5\% noise}
    \end{overpic}
    \caption{\textbf{Completion under varying noise levels}. Given an input with increasing noise levels, our method completes shapes plausibly with a consistent global structure and geometry. On the other hand, minor artifacts can be observed at the highest noise level.}
    \label{fig:noise_robustness}
\end{figure}

\begin{figure}[!t]
    \centering
    \begin{overpic}[width=0.95\linewidth]{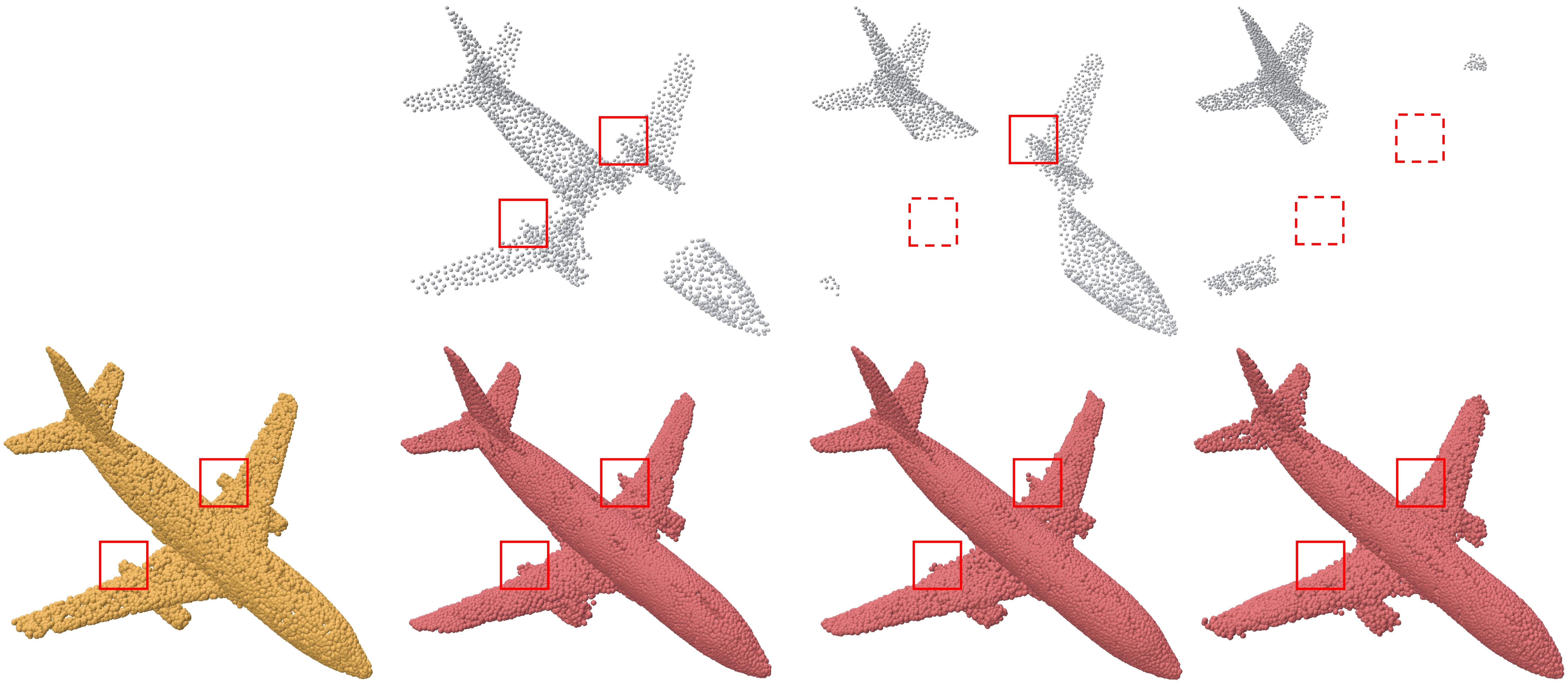}
    \put(10, -3){\small GT}
    \put(28.8, -3){\small 25\% missing}
    \put(54.9, -3){\small 50\% missing}
    \put(79.3, -3){\small 75\% missing}
    \end{overpic}
    \caption{\textbf{Completion under varying missing ratios}. Given an input partial shape with varying missing ratios, our method robustly reconstructs the shape. 
    }
    \label{fig:ratio_robustness}
\end{figure}

\paragraph{Robustness to noise.}
Fig.~\ref{fig:noise_robustness} visualizes \mname under increasing input noise. To simulate sensor noise, we add zero-mean Gaussian perturbations to the coordinates of partial point clouds, with standard deviations of $0.5\%$, $1.0\%$, and $1.5\%$ relative to the object scale. Across all settings, \mname preserves the overall structure, and even at $1.5\%$ noise it reconstructs the aircraft with correct global extent and coherent geometry. This robustness stems from the multi-view prior: the VLM-completed views provide consistent global evidence that stabilizes prior acquisition and guides the coarse-to-fine network beyond local coordinate jitter.

\paragraph{Robustness to missing ratios}
Fig.~\ref{fig:ratio_robustness} visualizes \mname under varying levels of structural incompleteness. To evaluate the performance across different degrees of data loss, we downsample the partial point clouds by 25\%, 50\%, and 75\%. Across all settings, \mname demonstrates consistent stability, reconstructing complete geometry with accurate proportions and coherent global extent.
As highlighted by the red boxes, the wing engines remain visible at 25\% and 50\% missing ratios, and \mname reconstructs them consistently with the ground truth. At 75\% missing, the wing engine evidence is largely absent, so the prediction can deviate from the GT while still remaining geometrically plausible. This indicates that \mname is strong across missing ratios, and under severe incompleteness it increasingly relies on the scale-consistent multi-view prior inferred from VLM-completed renderings.

\begin{figure}[!t]
    \centering
    \begin{overpic}[width=0.95\linewidth]{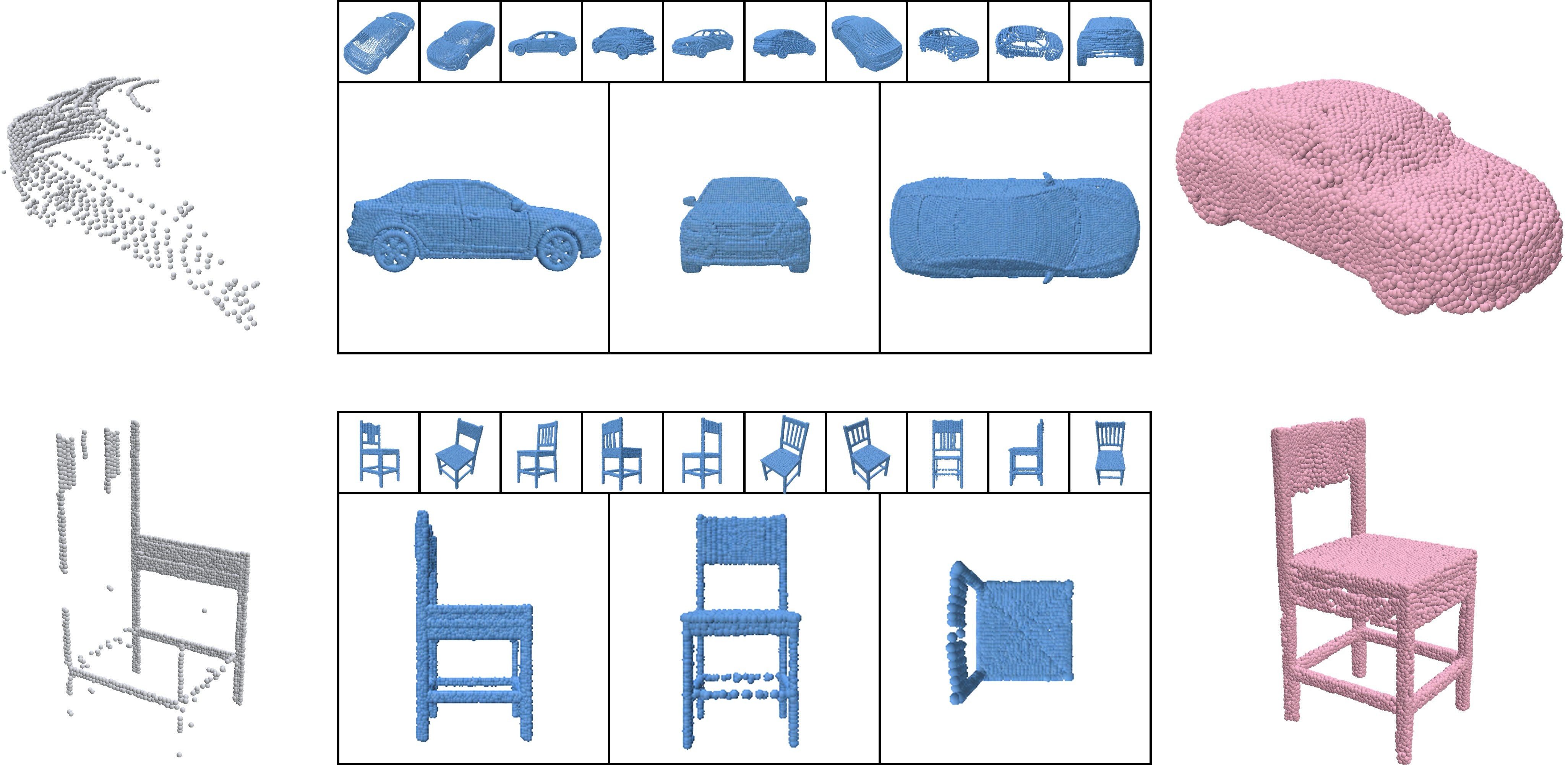} 
    \put(4, -3.5){\small (a) Input}
    \put(31, -3.5){\small (b) VLM output images}
    \put(81, -3.5){\small (c) Result}
    \end{overpic}
    \caption{\textbf{LiDAR point completion}. Given the partial LiDAR point cloud with a large proportion of missing regions, our method can produce structurally valid and geometrically plausible completion.}
    \label{fig:lidar}
\end{figure}

\paragraph{LiDAR point completion.}
Figure~\ref{fig:lidar} presents qualitative results on challenging real-world datasets---indoor scenes from ScanNet~\cite{dai2017scannet} and outdoor driving scenes from KITTI~\cite{geiger2013vision}. 
These sensor observations are typically sparse and noisy (see Fig.~\ref{fig:lidar}(a)).
However, the pretrained VLMs can still leverage their learned shape and semantic knowledge to generate high-quality multi-view 2D prior images (Fig.~\ref{fig:lidar}(b)).
The final results (Fig~\ref{fig:lidar}(c)) indicate that our network effectively utilizes these 2D priors as guidance to recover complete, geometrically consistent 3D shapes without further finetuning on specific datasets. 
This experiment demonstrates the generalization capability and potential for practical applications of our approach.

\begin{figure}[!t]
    \centering
    \begin{overpic}[width=0.99\linewidth]{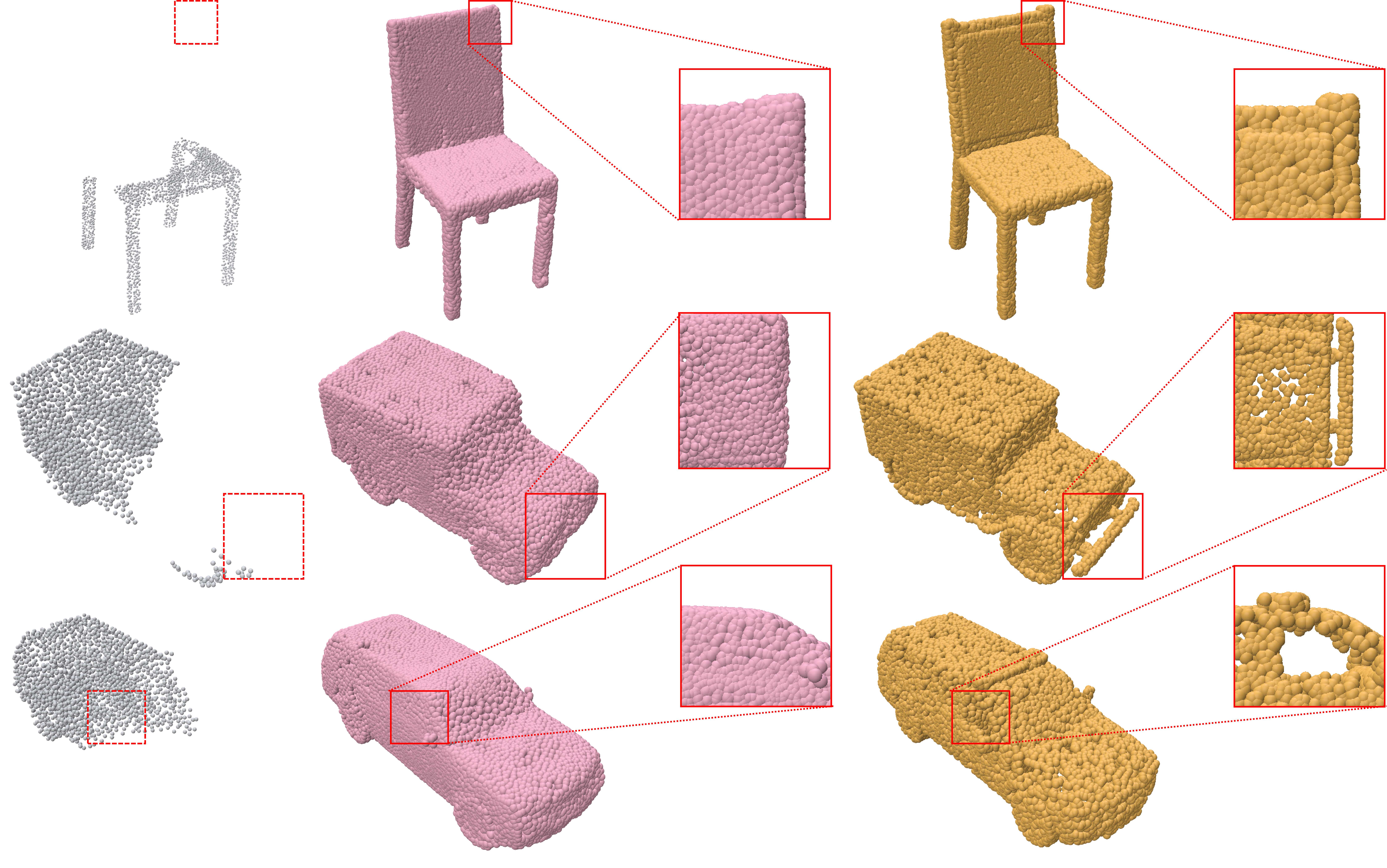} 
    \put(4.5, -3){\small Input}
    \put(42.5, -3){\small Ours}
    \put(84, -3){\small GT}
    \end{overpic}
    \caption{\textbf{Qualitative results showing plausible variations from the ground truth.} When dealing with unknown structures, our method synthesizes geometrically reasonable alternatives rather than producing results completely identical to the exact GT.}
  \label{fig:different_result}
\end{figure}

\paragraph{Plausibility versus overfitting.} 
Furthermore, the exact unobserved geometry cannot always be deterministically recovered from a highly incomplete partial scan. Rather than simply memorizing the ground-truth shapes---a common symptom of overfitting---our network learns to hallucinate geometrically plausible and structurally sound completions. As illustrated in Fig.~\ref{fig:different_result}, our results may occasionally exhibit slight structural variations from the exact GT in severely occluded regions. For instance, the generated edges of the chair (first row) and the joint structure between the backrest and seat cushion (second row) differ subtly from the original GT. Similarly, natural variations can be observed in the generated car windows (third row) and bumpers (fourth row). These discrepancies are highly reasonable and expected; they provide evidence that our model can generalize to synthesize valid geometric alternatives based on robust priors, rather than rigidly overfitting to the specific local details of the training data.

\begin{figure}[!t]
    \centering
    \begin{overpic}[width=0.99\linewidth]{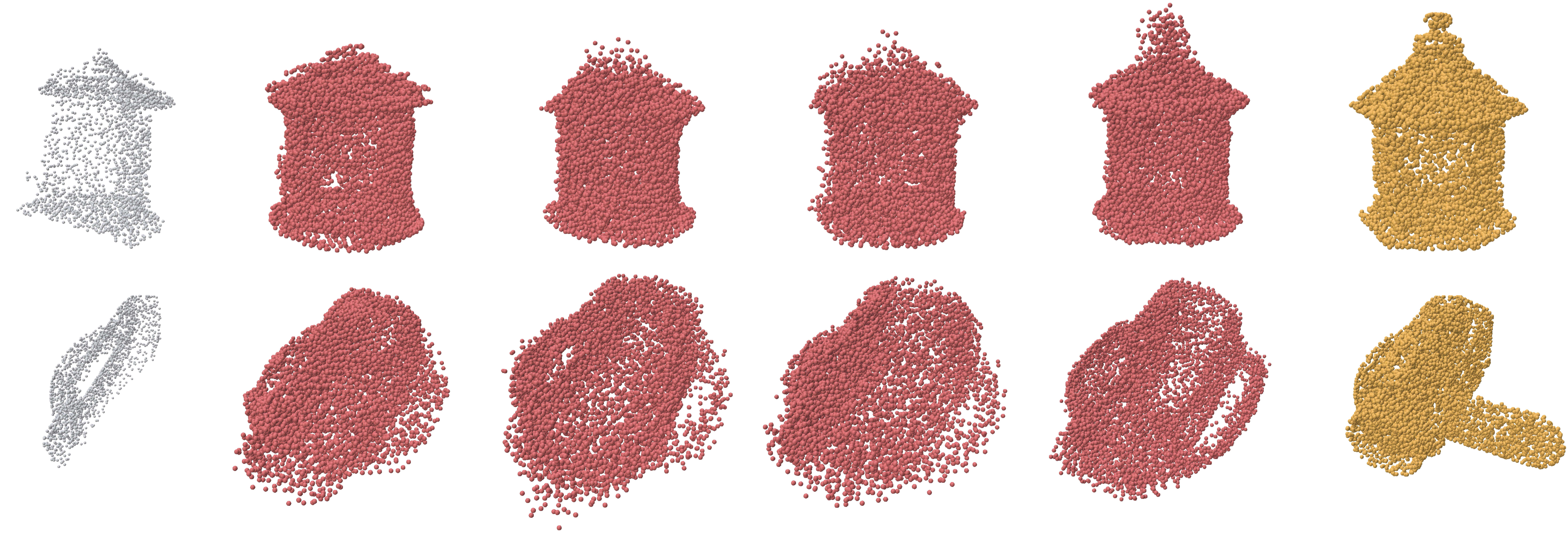} 
    \put(3, -3){\small Input}
    \put(15, -3){\small AdaPoinTr}
    \put(32, -3){\small GeoFormer}
    \put(49, -3){\small SVDFormer}
    \put(67, -3){\small PCDreamer}
    \put(90, -3){\small GT}
    \end{overpic}
    \caption{\textbf{Qualitative results of SoTA methods under the partial-based normalization setting.} 
    When baseline methods are forced to normalize inputs using the observed partial bounding box instead of the ground truth, they suffer from severe geometric distortion and scale collapse.}
  \label{fig:partial_normalization}
\end{figure}
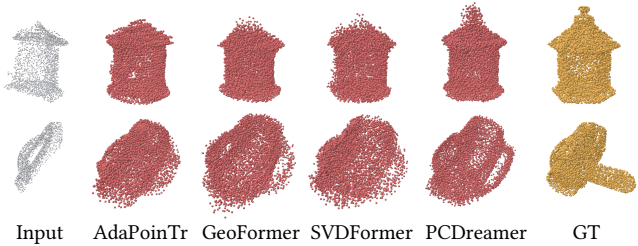

\paragraph{On-the-fly partial-based normalization.}
This setting is practical, as it normalizes the input partial point cloud solely based on the observation itself, without accessing any scale information from the full shape.
To evaluate existing completion methods under this realistic protocol, we normalize the partial input on-the-fly and directly feed it into their networks.
However, most prior methods implicitly assume GT scale during both training and inference through GT-scale-tied canonicalization, resulting in a significant distribution shift once this oracle cue is removed.
Consequently, naively applying their pretrained models under partial-only normalization yields unreasonable completions, as shown in Fig.~\ref{fig:partial_normalization}.
As expected, the performance degrades substantially.
Without access to full-shape scale (either from GT or reliable estimation), completion becomes considerably harder, as the model must jointly infer the global placement/scale of the partial observation and hallucinate the missing geometry.
These results further validate the motivation of our method, which explicitly recovers the missing global extent and scale under the GT-scale-free setting.

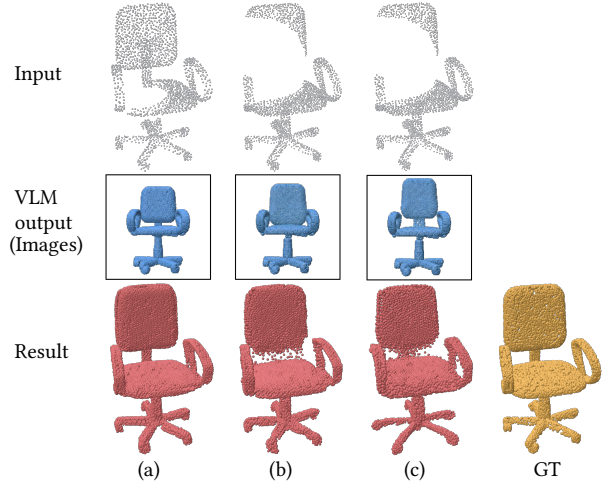
\begin{figure}[!t]
    \centering
    \begin{overpic}[width=0.85\linewidth]{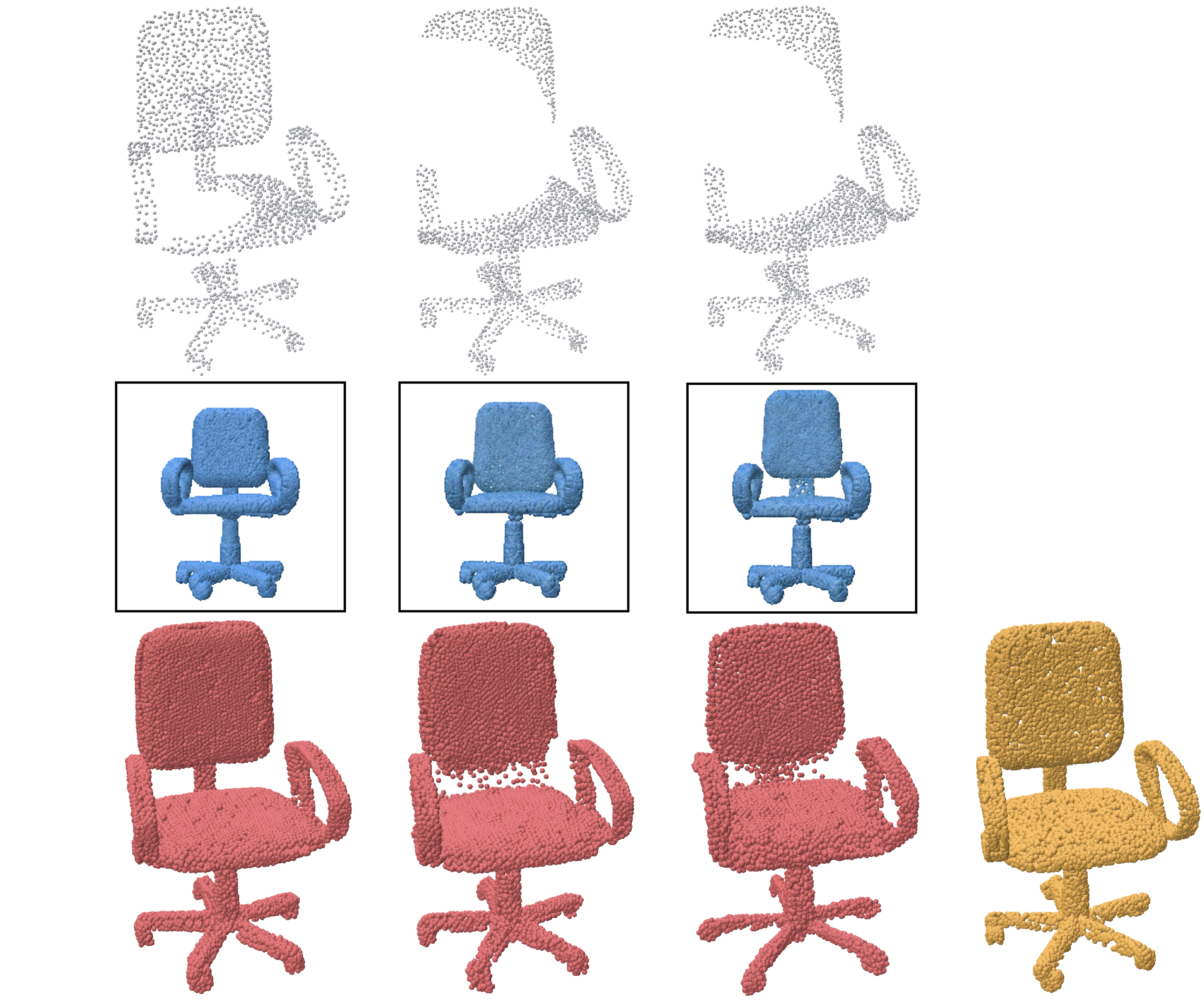} 
    \put(87.5, -3.5){\small GT}
    \put(15.4, -3.5){\small (a)}
    \put(39.3, -3.5){\small (b)}
    \put(63.9, -3.5){\small (c)}
    \put(-7, 68.5){\small Input}
    \put(-7, 18){\small Result}
    \put(-7, 46){\small VLM}
    \put(-7, 41.5){\small output}
    \put(-8, 37.5){\small (Images)}
    \end{overpic}
    \caption{\textbf{VLM inpainting and 3D shape completion.}
    Columns (a)–(c) show completion results conditioned on different VLM-inpainted images. Due to varying visibility of the joint between the backrest and the seat in the partial input point cloud, VLMs may hallucinate different geometric details, leading to distinct 3D completions.}
    \label{fig:failure_case}
\end{figure}

\paragraph{Failure case.}
Figure~\ref{fig:failure_case} visually illustrates the decisive influence of the 2D images hallucinated by the VLM on our final 3D completion results. 
We analyze this dependency by examining the recovery of the specific connection structure between the chair's backrest and seat cushion under different input conditions.
In Fig.~\ref{fig:failure_case}(a), geometric cues of the connecting structure remain partially visible in the input point cloud. Benefiting from these hints, the VLM successfully generates accurate and plausible 2D images, which in turn guide the model to produce high-quality 3D completion results that closely match the ground truth. Conversely, in Fig.~\ref{fig:failure_case}(b), this connecting structure is entirely missing from the partial input. Lacking local geometric constraints, the VLM hallucinates an alternative, incorrect structural arrangement. Consequently, the final 3D reconstruction faithfully follows this erroneous guidance from the VLM prior, resulting in a less ideal geometric shape that deviates from the original object. In Fig.~\ref{fig:failure_case}(c), although the connection is also completely absent in the input, the VLM utilizes its learned semantic priors to hallucinate a plausible (even if not identical to ground truth) connecting structure. As a result, our model reconstructs a 3D shape that is geometrically consistent with the VLM-provided image. 
Note that the noisy points in (b) and (c) may stem from inconsistencies in the VLM-inpainted images in the connecting region.

\paragraph{Limitation.}
From outer boundary cues, the rendered images successfully guide the VLM to generate accurate global profiles; however, lacking explicit depth or internal topological constraints, these 2D renderings cannot adequately capture depressions, indentations, or local concave surfaces.  Note that this performance bottleneck in handling local concavities is a direct consequence of the fundamental ambiguity of representing rich 3D internal geometries through flat 2D renderings.
Furthermore, the efficacy of our framework currently relies on the capabilities of \emph{closed-source} VLMs. Through extensive evaluations across various open- and closed-source models (\eg, FLUX\cite{labs2025flux1kontextflowmatching} and GPT-Image2), we observed that existing open-source models still struggle to produce view-consistent images, a property that is crucial for robust 3D lifting. In contrast, state-of-the-art proprietary models like Nano-banana~\cite{team2023gemini}, GPT-4o, and GPT-Image2 demonstrate the requisite multi-view consistency and high image quality. We hope that in the near future, new open-source models can serve our purpose.

\section{Conclusion}
We propose \mname, a novel point cloud completion framework that achieves breakthroughs on two parallel fronts: establishing a highly effective state-of-the-art 3D completion architecture, and solving the practical GT-scale-free challenge.
Our method leverages multi-view rendering and foundation-model-based image completion to infer missing extents, lifts the multi-view evidence into a confidence-aware 3D proxy, and fuses it with the observed partial input via a refinement network, which utilizes a dual-mode fusion strategy to meticulously align the 3D partial geometry with hierarchical 2D visual cues to generate the final completion.
Extensive experiments demonstrate that \mname consistently outperforms state-of-the-art completion methods and modern 3D generative baselines under realistic GT-scale-free settings. 
We hope our work can inspire further efforts on bridging foundation-model priors with 3D shape modeling and completion.

\paragraph{Future work.} There are a few inspiring directions worth exploring in the future. Our current framework assumes rotational alignment and does not explicitly handle rotational misalignment. 
While rotation-invariant point cloud encoders exist, their performance remains limited in practice; developing more effective rotation-robust representations is an important direction. 
The multi-view image priors provided by current foundation models are not always perfect, especially under severe occlusions and ambiguous structures. 
Future work may explore stronger multi-view consistency constraints and more reliable mechanisms for lifting 2D priors into 3D.


\bibliographystyle{ACM-Reference-Format}
\bibliography{main}

\clearpage
\section*{Supplemental Material}

\appendix

\setcounter{table}{0}
\renewcommand{\thetable}{A\arabic{table}}
\setcounter{figure}{0}
\renewcommand{\thefigure}{A\arabic{figure}}

In this supplementary material, we provide additional details on the data construction procedure, the prompt when using Nanobanana, more discussion, and visual comparison results on the ShapeNet-55 dataset.

\section{Dataset Construction Details}

\begin{figure}[!t]
    \centering
    \begin{overpic}[width=\linewidth]{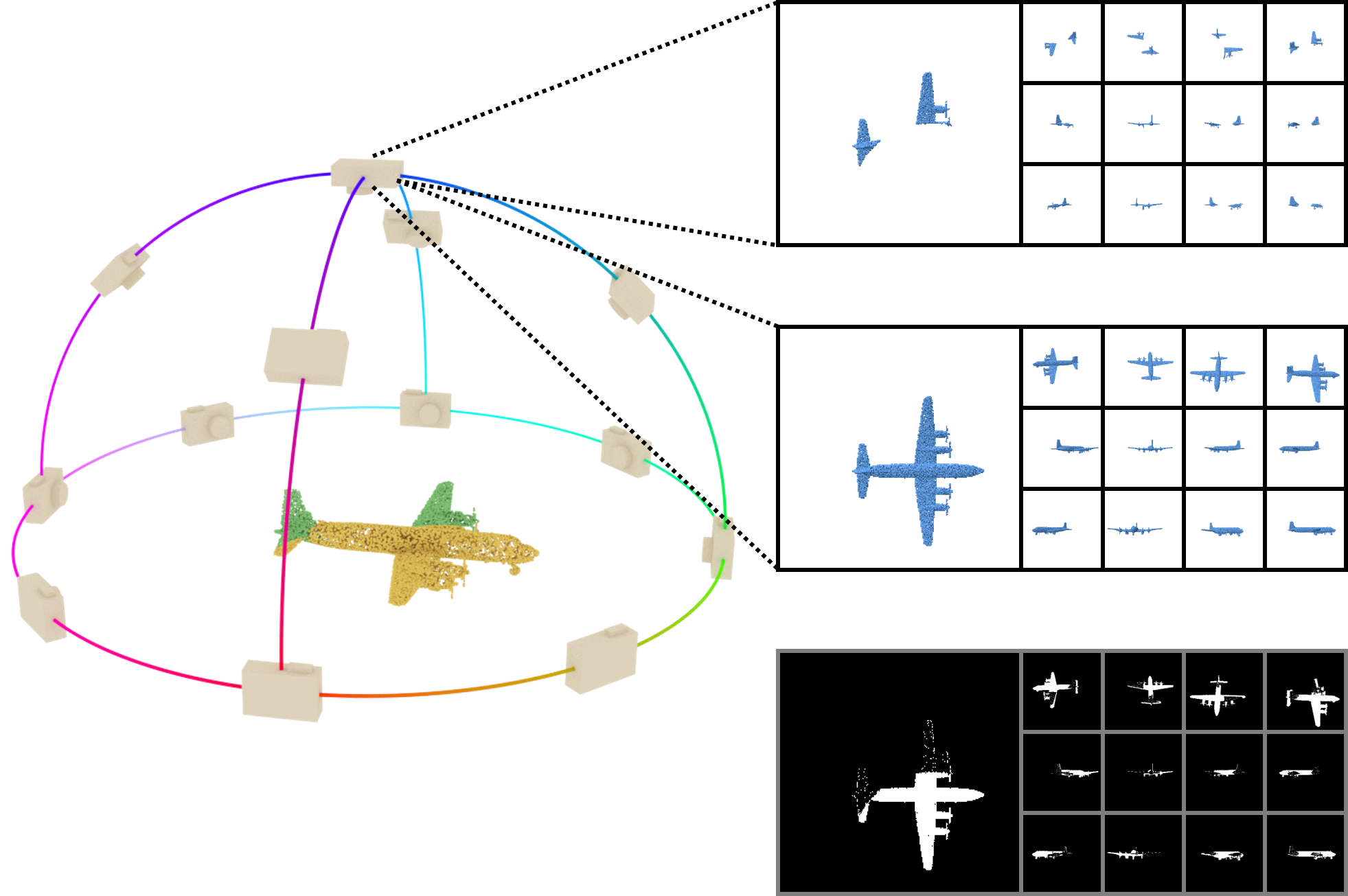} 
    \put(16, 8){\small Multi View Projection}
    \put(61.5, 45){\small Partial Projected Images}
    \put(59.5, 21){\small Complete Projected Images}
    \put(58, -3){\small Pixel-wise Differencing Masks}
    \end{overpic}
    \caption{\textbf{Dataset construction.}
    Partial projections are rendered from $N_v$ canonical views of the input point cloud and paired with completed projections (VLM at inference; simulated during training). Hallucination masks are obtained by pixel-wise differencing between the partial and completed projections.}
  \label{fig:dataset_variation}
\end{figure}

Figure~\ref{fig:dataset_variation} demonstrates a few data processing steps of the construction.
Specifically, for each shape from the benchmark dataset, we first create the partial observation (green) and the removed points (yellow) following the strategy as described in Sec. 4 in the main paper. 
Intuitively, from the point partitions, we could follow the steps as in our prior acquisition module to generate the training samples. 
However, constructing training data by invoking a VLM to complete multi-view projections for every sample is prohibitively expensive. To make construction practical while staying consistent with our inference pipeline, we adopt a lightweight data simulation procedure that mimics VLM-based multi-view completion and injects realistic view-dependent uncertainty.

The key observation is that VLM-completed projections rarely corrupt the underlying object geometry, based on extensive preliminary experiments.
Instead, their imperfections mainly arise from deviations in camera viewpoints, misalignment across views, and mild scale drift.
Motivated by this observation, we simulate VLM uncertainty by perturbing camera parameters when projecting the geometry.
This design preserves the geometric integrity of the object while producing projection artifacts that are consistent with those observed during inference.

Specifically, as shown in Fig.~\ref{fig:dataset_variation}(left), we directly project the partial and the full points from a set of predefined canonical viewpoints, generating the multiview partial and completed projected images.
For each view, random perturbations are applied to both the camera position and the look-at center.
For most viewpoints, small Gaussian noise is added to the camera position with a standard deviation of $0.06$, and to the look-at center position with a standard deviation of $0.03$, simulating mild viewpoint misalignment.
For a subset of challenging viewpoints (\eg the top views), stronger perturbations are applied, with camera position noise of standard deviation $0.15$ and target noise of $0.06$.
In addition, mild scale variations are introduced along the viewing direction for these views, with a random scaling factor sampled from $[0.7, 0.9]$, mimicking the scale drift commonly observed in VLM completions.
The two multi-view image sets support the calculation of the pixel-difference masks, from where we apply the remaining steps as in the prior acquisition procedure, including voxel voting, confidence-based shape fusion, and joint normalization (see Fig. 3 in the main paper).

After the whole process, for each data sample, we have the paired \{\emph{partial points}, \emph{multiview full images}, \emph{ground truth full points}\} for training and testing.

\section{Implementation Details}

\paragraph{Image inpainting prompt}
In the following, we show the inpainting prompt when interacting with Nanobanana.\\

\begin{minipage}{\columnwidth}
\begin{lstlisting}[label={lst:nano_banana_prompt}]
Input: a rendering of a partial point cloud observation.

Task_Prompt:

Task Objective: Based on a visualization of an incomplete chair point cloud, generate a structurally complete visualization of that same point cloud.

Viewpoint Specification: the output image must strictly adhere to this exact viewpoint and camera angle.

General Principles:
-Analyze Valid Information: Comprehensively analyze the geometry, object contours, and spatial distribution.
-Infer Complete Stucture: Infer the most probable geometry of the missing parts.
-Smooth Continuity: Ensure smooth and natural transitions.

Mandatory Constraints:
-Rough Shape Completion: Only complete the approximate overall shape.
-Avoid Excessive Detail: Do not reconstruct fine textures or minor sub-components.
-Prohibit Invention: Do not invent structures not implied by the data.

The output Format and Content:
-Format Consistency: the output style and format should be identical to the input.
-Pure Content: Completed object only, white background.

Output: 
An inpainted image depicting the full shape.

\end{lstlisting}
\end{minipage}

\end{document}